\documentclass{article}

\usepackage{color}
\usepackage{empheq}

\usepackage{geometry}

\usepackage{float}
\usepackage{subfiles}
\usepackage{xr}
\usepackage{amsmath,amssymb,amsfonts}
\usepackage{mathtools}
\usepackage{mathrsfs}
\usepackage{makeidx}
\usepackage{graphicx}
\usepackage{cases}

\usepackage{exscale,relsize} 

\usepackage{mleftright}
\usepackage{theorem}
\usepackage{graphics}                 
\usepackage{accents}
\usepackage{kbordermatrix}
\usepackage[margin=10pt,font=small,labelfont=bf]{caption}
\usepackage{tikz}
\usepackage{subfigure}
\tikzstyle{every picture}=[scale=0.75,transform shape]
\usepackage{verbatim}
\usetikzlibrary{arrows,automata}

\usepackage{bm}
\usepackage{longtable}
\usepackage{tabu}
\usepackage{algorithmic}
\usepackage{algorithm}

\usepackage{url}

\usepackage{bm,xstring}

\def\expm{\mathop{\mathrm{expm}}}

\newcommand{\sparsemin}{\mathop{\mathrm{spmin}}}
\newcommand{\minimize}{\mathop{\mathrm{minimize}}}
\newcommand{\maximize}{\mathop{\mathrm{maximize}}}
\newcommand{\subjectto}{\mathop{\mathrm{subject\,to}}}

\newcommand{\myDelta}{{\textstyle \mathsmaller{\varDelta}}} 

\newcommand{\dsum}{\displaystyle\sum}

\newcommand{\cditto}{\raisebox{.5ex}{\hbox to 2em{\hrulefill}}}

\newcommand{\bigzero}{\mbox{\normalfont\large\bfseries 0}}

\mleftright 

\title{Sparse Randomized Shortest Paths Routing\\ with Tsallis Divergence Regularization\\
\quad \\
  \normalsize{\textit{Draft manuscript subject to changes}}}

\author{Pierre Leleux, Sylvain Courtain, Guillaume Guex \& Marco Saerens}
\begin{document}
\setlength{\parskip}{1pt plus 1pt minus 1pt} 

\sloppy 

\date{}
\maketitle
\begin{abstract}
This work elaborates on the important problem of (1) designing optimal randomized routing policies for reaching a target node $t$ from a source note $s$ on a weighted directed graph $G$ and (2) defining distance measures between nodes interpolating between the least cost (based on optimal movements) and the commute-cost (based on a random walk on $G$), depending on a temperature parameter $T$. To this end, the randomized shortest path formalism (RSP, \cite{Akamatsu-1996,Saerens-2008,Yen-08K}) is rephrased in terms of Tsallis divergence regularization, instead of Kullback-Leibler divergence. The main consequence of this change is that the resulting routing policy (local transition probabilities) becomes sparser when $T$ decreases, therefore inducing a sparse random walk on $G$ converging to the least-cost directed acyclic graph when $T \rightarrow 0^{+}$. Experimental comparisons on node clustering and semi-supervised classification tasks show that the derived dissimilarity measures based on expected routing costs provide state-of-the-art results. The sparse RSP is therefore a promising model of movements on a graph, balancing sparse exploitation and exploration in an optimal way.
\end{abstract}

\section{Introduction}
\label{Sec_introduction01}

\subsection{General introduction}

Link analysis and network science are currently used in a large number of different fields for analysing network data, like social networks, networks of transactions, protein networks, road networks, etc (see, e.g., \cite{Barabasi-2015,Brandes-2005,chung06,Estrada-2012,Fouss-2016,Kolaczyk-2009,Lewis-2009,Newman-2018,Silva-2016,Thelwall04,Wasserman-1994}).
In this context, many important problems arise, such as the definition of meaningful distance measures between nodes or models of movement / communication in the network (routing policies). These quantities usually take the structure of the whole network into account.

One such model is the \textbf{randomized shortest paths} (\textbf{RSP}; see the related work), first developed in transportation sciences \cite{Akamatsu-1996} and then further extended in the context of network data analysis \cite{Francoisse-2017,Kivimaki-2012,Saerens-2008,Yen-08K}. Assume an agent walking on the graph who wants to reach a target node $t$ from some source node $s$ by following some path or walk connecting these two nodes. The RSP uses a statistical physics formalism by putting a Gibbs-Boltzmann probability distribution on the set of paths\footnote{Considered as independent.} from $s$ to $t$, depending on a temperature parameter $T$. The agent then chooses a path to follow according to this Gibbs-Boltzmann distribution -- a \emph{global policy} or \emph{routing strategy} in terms of paths. When $T$ is low, low-cost paths are favored while when $T$ is large, paths are chosen according to their likelihood in a completely random walk on the graph $G$. It has been shown that this model of movement defines, at the local node level, a \emph{biased random walk} on $G$ attracting the walker to the target node. This biased random walk defines the \emph{local policy} captured by the transition probabilities matrix of an absorbing Markov chain. The model can be derived by minimizing expected path cost regularized by Kullback-Leibler (\textbf{KL}) divergence based on path probabilities, and thus provides an optimal policy in that sense (see the above references for details).

Inspired by \cite{Akamatsu-1997,Bavaud-2012,Saerens-2008}, this work first shows how the paths-based minimization problem can be rephrased in terms of a local objective function, that is, the cost function and the regularization term are expressed in terms of transition probabilities instead of path probabilities. This reformulation allows to replace the KL divergence by the Tsallis divergence (\cite{Tsallis-1998,Tsallis-2009}, also called Havrda-Charvat divergence \cite{Havrda-1967,Kapur-1989}), with the consequence that the obtained policy becomes \emph{sparse} for low temperatures (as, e.g., in \cite{Li-2011,Li-2013}; see the related work). In these conditions, the resulting biased random walk is only defined on a subset of edges (a subgraph of $G$) and when $T$ decreases toward zero, the most costly paths are gradually removed until only shortest paths remain active (see the illustration on Figure \ref{fig:NetFlows}). More generally, the model of movement interpolates between a pure random walk (large $T$, exploration) and the least cost paths (low $T$, exploitation). The proposed algorithm iteratively computing the policy is inspired by \cite{Kanzawa-2013,Miyamoto-1998,Saerens-2008} and solves the primal problem, providing transition probabilities, and then the dual problem for computing the Lagrange parameters which correspond to the minimized free energy, until convergence.

In addition, two new \emph{dissimilarity measures} between nodes are derived from this framework, the Tsallis RSP and \textbf{free energy} (\textbf{FE}) dissimilarities\footnote{We conjecture that the Tsallis FE is a distance measure as it verified the properties of a distance measure in all our applications.}. These are the counterparts of the same dissimilarities defined in the standard RSP framework based on the KL divergence: the RSP is the expected cost from $s$ to $t$ when following the optimal randomized policy and the FE is the minimized free energy objective function at the optimal routing policy \cite{Kivimaki-2012}. Both capture the notion of \emph{relative accessibility} (proximity and amount of inter-connectivity \cite{Chebotarev-1998a}) between nodes and interpolate (up to a scaling factor) between the least-cost and the commute-cost distances on undirected graphs.

\subsection{Related work}
\label{Subsec_related_work01}

Traditional network measures are usually derived from two different paradigms about movement, or communication, taking place in the network \cite{Guex-2019}: optimal communication based on least cost paths, and random communication based on a random walk on the graph.
For instance, the shortest path distance and the standard betweenness centrality \cite{Freeman-1977} are defined from least cost paths whereas resistance distance and random walk centrality \cite{Brandes-2005b,Newman-05} rely on random walks \cite{Snell-1984}.
But, in practice, movements over a network hardly ever occur either perfectly optimally or perfectly randomly: the agent usually has some, although incomplete, knowledge of his environment.
As already mentioned, the RSP framework \cite{Francoisse-2017,Kivimaki-2012,Saerens-2008,Yen-08K} relaxes these assumptions by interpolating between least cost paths and a pure random walk on the graph, depending on a temperature parameter. In this context, the RSP has been used recently, e.g., for modelling the behavior of animals during migration \cite{Panzacchi-2016}. Initially, the model was inspired by transportation models developed in transportation sciences \cite{Akamatsu-1996}, and is also closely related to the work of \cite{Todorov-2007,Todorov-2008} in reinforcement learning and \cite{Delvenne-2011} defining random walks on a graph with minimal free energy or maximal entropy.

Besides these previous works, other interesting models of movement interpolating between random and optimal behaviors have been proposed in the recent literature. Some of these models are based on electrical networks and extensions \cite{vonLuxburg-2011,Herbster-2009,Nguyen-2016,Luxburg-2014}, others on combinatorial analysis arguments \cite{Chebotarev-2011,Chebotarev-2012,Chebotarev-2013}, on L1-L2 regularization \cite{Li-2011,Li-2013}, as well as on  network flow models with Kullback-Leibler divergence regularization \cite{Bavaud-2012,Guex-2015}. These last propositions are equivalent to the RSP, although derived from another perspective.
Many of these models lead to distance measures between nodes avoiding the so-called ``lost in space" effect \cite{Luxburg-2010,Luxburg-2014} by interpolating between the least cost distance and the resistance distance (equivalent to the commute-time distance and the commute-cost distance, up to a scaling factor \cite{FoussKDE-2005}).
For a more thorough discussion of these related works, see, e.g., \cite{Fouss-2016,Francoisse-2017}.

Among these references, the most closely related works are \cite{Li-2011,Li-2013}. The authors present a sparse routing strategy based on edge flow optimization regularized by mixed L1/L2 norms \cite{Hastie-2015}. More precisely, they use the edge flow formalism \cite{Ahuja-1993,Dolan-1993} and minimize the weighted sum of (non-negative) net flows and squared net flows subject to flow conservation constraints.
Although these papers address problems similar to our work and are of high interest, there are significant differences. First, they consider undirected networks while we address directed ones. They also consider net flows (like in electrical networks) while the present work considers raw flows. Moreover, the regularization technique is different: in \cite{Li-2011,Li-2013}, a mixed L1/L2 norm is minimized whereas we use a Tsallis divergence regularization term\footnote{See the discussion following Equation (\ref{Eq_Tsallis_formulation_lagrange_function01}) for details about the differences between the two objective functions.}. Therefore, the derived algorithms are different.
In addition, our work extends the scope of the RSP framework by using the Tsallis divergence instead of the KL divergence, whereas the \cite{Li-2011,Li-2013} work is based on network flow theory. Furthermore, one of our main goals is to derive dissimilarity measures between nodes, which is not addressed in \cite{Li-2011,Li-2013}.

Concerning Tsallis entropy regularization, it is known for years that when minimizing the regularized expected cost together with sum-to-one and non-negativity constraints, the resulting solution becomes sparse.
This is not surprising because the Tsallis entropy, in its basic form, is the L2 squared norm of a non-negative vector whose entries sum to one (a probability mass). It is therefore closely related to L1/L2 regularization which is known to provide sparse solutions \cite{Buhlmann-2011,Hastie-2009,Hastie-2015}.
For instance, it was shown in \cite{Kanzawa-2013,Miyamoto-1998} that this technique provides sparse discrete cluster membership probabilities in fuzzy clustering. In this context, a discussion of the use of various regularization terms in fuzzy clustering is provided in \cite{Menard-2003,Miyamoto-2008}; however, \cite{Menard-2003} considers a different objective function which does not lead to sparse solutions. Another interesting application is the sparse probabilistic latent semantic analysis introduced in \cite{Hazan-2007}. The author proposes an expectation-maximization algorithm based on Tsallis divergence, instead of KL divergence. In another paper \cite{Hazan-2007b}, the same authors study maximum Tsallis entropy estimation and propose an algorithm solving the problem. Interestingly, the paper also discusses the different ways of defining maximum Tsallis entropy (see \cite{Hazan-2007b}, subsection 1.2), with only one of these ways being relevant in our applications.
Note also that in \cite{Kanzawa-2018}, Kanzawa considers Tsallis divergence regularization, but again uses a different objective function that does not lead to a sparse solution.

Also closely related to our work, the recent interesting paper \cite{Lee-2018} investigates sparse policies for discounted Markov decision processes (\textbf{MDP}), with application to reinforcement learning (see also \cite{Geist-2019}). As initiated in \cite{Todorov-2007,Todorov-2008} (and related to the alternative view of the RSP presented in Subsection \ref{Subsec_alternative_randomized_shortest_paths01}), the authors add an entropy term to the cost associated to each action choice in the MDP. Instead of using the KL divergence as in \cite{Todorov-2007}, they choose the Tsallis entropy with $r=2$ and a uniform reference probability distribution (see Equation (\ref{Eq_Tsallis_r_divergence01}) below). As in \cite{Kanzawa-2013,Miyamoto-1998}, they show that the resulting policy enforces sparsity. This problem is in fact closely connected to the recently introduced sparse maximum procedure \cite{Laha-2018,Martins-2016} and the probability simplex projection, e.g. \cite{Beck-2014,Condat-2016,Duchi-2008,Wang-2013b} who proposed procedures similar to \cite{Kanzawa-2013,Miyamoto-1998} for solving this problem. Our work therefore solves a problem similar to \cite{Lee-2018} because a Markov chain is a particular case of a MDP. However, our results are complementary, as the optimal policy is derived from a different perspective (the RSP). Moreover, they extend to general Tsallis divergences of the form provided by Equation (\ref{Eq_Tsallis_r_divergence01}), with $r>1$ (generalizing Tsallis entropy), and thus also integrating a reference distribution. Therefore, these new results could easily be applied to the MDP problem, which will be considered in further work. Note also that we are working with absorbing Markov chains while \cite{Lee-2018} investigates a discounted process, but this is mainly a detail. Finally, as already stressed, our objective, in addition to derive optimal routing policies, is to introduce new dissimilarity measures interpolating between the least-cost and the commute-cost distances between nodes of a graph.

Other recent works related to the induced sparsity of Tsallis regularization are \cite{Martins-2016} for obtaining sparse outputs of a classifier in the context of multi-label classification \cite{Lee-2018b}, extending \cite{Lee-2018} and proposing a ``maximum causal Tsallis entropy" framework (see \cite{Lee-2018} for details). Finally, in \cite{Muzellec-2017}, the authors unify the two main approaches to optimal transport, namely Monge-Kantorovitch and Sinkhorn-Cuturi, into what they define the Tsallis-regularized optimal transport. As the name indicates, the objective function is regularized by Tsallis $r$-entropy and they then analyse the optimization problem by using the $r$-exponential formalism \cite{Tsallis-2009}. Then, a different approach from the one introduced in this paper is used in order to solve the problem, namely a sophisticated gradient-based algorithm.

\subsection{Contributions and contents of the paper}

In short, the main contributions of this paper are
\begin{itemize}
  \item The development and the investigation of a new randomized shortest paths model of movement on a graph, considering Tsallis divergence instead of the KL divergence. This model shows some interesting properties: it provides sparse routing policies and it interpolates between least-cost and random-walk routing, depending on the temperature parameter $T$.
  \item A procedure for computing the optimal policy (sparse when $T$ is small), minimizing expected cost plus Tsallis divergence, is derived. It therefore extends previous results by minimizing the Tsallis divergence instead of the Tsallis entropy.
  \item Two new dissimilarity measures between nodes, interpolating\footnote{For an undirected graph and up to a scaling factor.} between the least cost and the commute-cost distances based on a model of sparse movements on the network are introduced.
  \item It provides an experimental comparison between the two new dissimilarities based on Tsallis divergence regularization and other baseline dissimilarities on nodes clustering and semi-supervised classification tasks.
\end{itemize}

The content of the paper is as follows. Section \ref{Sec_standard_randomized_shortest_paths01} provides a brief introduction to the  RSP framework and introduces an alternative view on the RSP which can easily be generalized by considering other entropic regularizations. Then, in Section \ref{Sec_sparse_randomized_shortest_paths01}, the sparse RSP model of movement is developed based on Tsallis divergence regularization. Section \ref{Sec_distances01} introduces the two new dissimilarity measures between nodes. Illustrative examples and experiments on node clustering and semi-supervised classification tasks are presented in Section \ref{Sec_experiments01}. Finally, Section \ref{Sec_conclusion01} concludes the work.

\section{The randomized shortest paths framework}
\label{Sec_standard_randomized_shortest_paths01}

This section provides a short summary of the standard randomized shortest paths framework, based on the KL divergence, before introducing the alternative formulation (Subsection \ref{Subsec_alternative_randomized_shortest_paths01}) and then developing its sparse version (Section \ref{Sec_sparse_randomized_shortest_paths01}).

\subsection{Background and notation}

Let us first introduce some background and notation \cite{Fouss-2016,Francoisse-2017}.
Consider a weighted directed\footnote{If the graph is undirected, it is assumed that each undirected edge is composed of two directed edges with the same weight in the two opposite directions (reciprocal edges).}, strongly connected, graph, $G = (\mathcal{V}, \mathcal{E})$, with a set $\mathcal{V}$ of $n$ nodes and a set $\mathcal{E}$ of directed edges. The directed edge connecting node $i$ to node $j$ is denoted by $(i,j)$.
Moreover, the \emph{adjacency matrix} $\mathbf{A} = (a_{ij}) \ge 0$ of $G$ contains directed affinities between nodes. We further assume that there are no self-loops in the network, that is, $a_{ii}=0$ for all $i$ (a simple graph).

A natural random walk on the graph is defined from this adjacency matrix in the usual way. The \emph{reference transition probabilities} associated to each node are set proportionally to the affinities of the incident edges and then normalized in order to sum to one. Alternatively, they can also be chosen as uniform, depending on the problem,
\begin{equation}
p_{ij}^{\mathrm{ref}} = \frac{a_{ij}}{{\sum_{j'=1}^{n}} a_{ij'}} \text{ \footnotesize{ (natural random walk) } } \text{ or} \quad p_{ij}^{\mathrm{ref}} = \frac{1}{|\mathrm{S}ucc(i)|} \text{ \footnotesize{ (uniform distribution) } }
\label{Eq_Transition_probabilities01}
\end{equation}
where $\mathrm{S}ucc(i)$ is the set of successor nodes of node $i$ and $| \mathrm{S}ucc(i) |$ its cardinality.
The matrix $\mathbf{P}_{\mathrm{ref}} = (p_{ij}^{\mathrm{ref}})$ is stochastic and is called the transition matrix of the reference random walk on the graph.

Furthermore, a transition \emph{cost}, $c_{ij} \ge 0$, is
associated to each edge $(i,j)$ of the network $G$ and the resulting cost matrix is defined as $\mathbf{C} = (c_{ij})$.
If there is no edge linking $i$ to $j$ ($a_{ij}=0)$, the cost is assumed to take a very large value and the product $a_{ij} c_{ij} = 0$ by convention. Usually, costs are set independently of the affinities, depending on the application, but, if there are no obvious costs associated to the problem, we can, e.g., set $c_{ij} = 1/a_{ij}$ as in electric networks (see \cite{Francoisse-2017} for a discussion).

A path $\wp$ is a finite sequence of transitions to adjacent nodes on $G$ (including cycles), initiated from a source node $s$ and stopping in some target node $t$. This target node is transformed into an \emph{absorbing} and \emph{killing} node. That is, when this node is reached, the random walkers stops his walk and disappears. This means that the corresponding row $t$ of the transition probabilities matrix $\mathbf{P}_{\mathrm{ref}}$ is set to zero -- the matrix therefore becomes sub-stochastic and represents a killed random walk on $G$.
The \emph{total cost} of a path, $\tilde{c}(\wp)$, is simply the sum of the edge costs $c_{ij}$ along $\wp$ while the \emph{length} of a path $\ell(\wp)$ (or simply $\ell$) is the number of steps needed for following that path from $s$ to $t$. Finally, the set of all paths connecting $s$ to $t$ is denoted by $\mathcal{P}_{st}$.

\subsection{The standard randomized shortest paths framework}
\label{Sec_randomized_shortest_paths01}

The \emph{randomized shortest paths} (RSP) framework \cite{Saerens-2008}, coming from transportation sciences \cite{Akamatsu-1996}, was further developed by exploiting basic concepts from statistical physics \cite{Francoisse-2017,Kivimaki-2012,Saerens-2008,Yen-08K}. A similar model was proposed in reinforcement learning and process control \cite{Todorov-2007,Todorov-2008}, from a different perspective. The RSP is based on a system of full paths (the ``macro" level -- a bag of paths connecting the source node to the target node\footnote{For a generalization to several input-outputs, see \cite{Guex-2019}.}) instead of standard ``local" flows defined on nodes and edges (the ``micro" level) \cite{Ahuja-1993,Dolan-1993}. Among others, it provides an optimal, randomized, routing policy from source node $s$ to target node $t$ based on the minimization over the set of paths $\wp \in \mathcal{P}_{st}$ of the (relative) KL-based free energy of statistical physics \cite{Jaynes-1957,Peliti-2011,Reichl-1998},
\begin{equation}
\vline\,\begin{array}{llll}
\minimize\limits_{\{ \mathrm{P}(\wp) \}_{\wp \in \mathcal{P}_{st}}} & \phi^{\mathrm{\textsc{kl}}}_{st}(\mathrm{P})
= \underbracket[0.5pt][3pt]{ \dsum_{\wp \in \mathcal{P}_{st}} \mathrm{P}(\wp) \, \tilde{c}(\wp)  }_{ \text{expected cost } }
+ T \underbracket[0.5pt][3pt]{ \dsum_{\wp \in \mathcal{P}_{st}} \mathrm{P}(\wp) \log \left( \frac{\mathrm{P}(\wp)}{\tilde{\pi}(\wp)} \right)  }_{ \text{KL  regularization } } \\[0.5cm]
\subjectto & \sum_{\wp\in\mathcal{P}_{st}}\textnormal{P}(\wp) = 1
\end{array}
\label{Eq_optimization_problem_BoP01}
\end{equation}
where $\tilde{c}(\wp) = \sum_{\tau = 1}^{\ell} c_{s(\tau-1) s(\tau)}$ is the total cumulated cost along path $\wp$ when visiting the sequence of nodes, or states, $\left( s(\tau) \right)_{\tau=0}^{\ell}$ in the sequential order and $\ell$ is the length of path $\wp$. Furthermore, $\tilde{\pi}(\wp) = \prod_{\tau = 1}^{\ell} p_{s(\tau-1) s(\tau)}^{\mathrm{ref}}$ is the random walk probability of the path, that is, the product of the reference transition probabilities (\ref{Eq_Transition_probabilities01}) along hitting path $\wp$ ending in (killing and absorbing) target node $t$.
%
%
The objective function is a mixture of two dissimilarity terms with the temperature $T>0$ balancing the trade-off between the two quantities.
The first term is the expected cost for reaching target node from source node (favoring shorter paths -- \emph{exploitation}). The second term corresponds to the Shannon relative entropy, or Kullback-Leibler divergence, between the path probability distribution and the reference path probability distribution (introducing randomness -- \emph{exploration}). For a low temperature $T$, shorter paths are favored whereas when $T$ is large, paths are chosen according to their probability in the reference random walk on $G$ (see Equation (\ref{Eq_Transition_probabilities01})).

As well-known, this free energy minimization problem, akin to maximum entropy models \cite{Cover-2006,Jaynes-1957,Kapur-1989}, leads to an \emph{Gibbs-Boltzmann} probability distribution on the set of paths (see, e.g., \cite{Francoisse-2017}),
\begin{equation}
\mathrm{P}^{*}(\wp) 
= \frac{\tilde{\pi}(\wp) \exp[-\theta \tilde{c}(\wp)]}{\dsum_{\wp'\in\mathcal{P}_{st}} \tilde{\pi} (\wp')\exp[-\theta \tilde{c}(\wp')]}
= \frac{\tilde{\pi}(\wp) \exp[-\theta \tilde{c}(\wp)]}{\mathcal{Z}}
\label{Eq_Boltzmann_probability_distribution01}
\end{equation}
where $\theta = 1/T$ is the inverse temperature and the denominator $\mathcal{Z} = \sum_{\wp\in\mathcal{P}_{st}} \tilde{\pi} (\wp)\exp[-\theta \tilde{c}(\wp)]$ is the \emph{partition function} of the system.
This provides the (randomized) \emph{optimal policy} in terms of paths to follow at the ``macro" (paths) level, that is, the \emph{optimal} probability distribution over all possible paths from $s$ to $t$.

It can further be shown that this probability distribution defines a local, optimal, policy which turns out to be a Markov chain at the ``micro" (node) level. In this context, the optimal transition probabilities of following any edge $(i,j)$ (the ``local policy") induced by the set of paths $\mathcal{P}_{st}$ and their probability mass (\ref{Eq_Boltzmann_probability_distribution01}) are\footnote{Here, we used $\phi^{\mathrm{\textsc{kl}}}_{it}(\mathrm{P}^{*}) = -\tfrac{1}{\theta} \log[z_{it}]$ and $p^{*}_{ij} = p_{ij}^{\mathrm{ref}} \exp[- \theta c_{ij}] z_{jt} / z_{it}$ (see \cite{Francoisse-2017,Kivimaki-2012,Saerens-2008} for details).} 
\begin{equation}
p^{*}_{ij} = p^{\mathrm{ref}}_{ij} \exp \left[ -\theta (c_{ij} + \phi^{\mathrm{\textsc{kl}}}_{jt}(\mathrm{P}^{*}) - \phi^{\mathrm{\textsc{kl}}}_{it}(\mathrm{P}^{*})) \right]
\text{ for all } i \ne t
\label{Eq_path_transition_probabilities01}
\end{equation}
and $p^{*}_{tj} = 0$ for target node $t$ and all $j$.
The free energy between $i$ and $t$ at the optimal probability distribution\footnote{Providing the directed free energy distance, see \cite{Kivimaki-2012}.}, $\phi^{\mathrm{\textsc{kl}}}_{it}(\mathrm{P}^{*})$, can be computed by solving a system of linear equations of size $n$ \cite{Francoisse-2017,Kivimaki-2012,Saerens-2008}.
As already mentioned, it defines a \emph{biased} random walk on $G$ where the random walker is more and more ``attracted" by the target node $t$ when $T$ decreases. Interestingly, the policy is independent of the source node $s$.
 It also implies that for any path $\wp$ visiting nodes $s(\tau)$, $\tau = 0, \dots, \ell(\wp)$, we have $\mathrm{P}^{*}(\wp) = \prod_{\tau = 1}^{\ell} p^{*}_{s(\tau-1) s(\tau)}$.

\subsection{An alternative form for the randomized shortest paths}
\label{Subsec_alternative_randomized_shortest_paths01}

The previous path-based objective function (\ref{Eq_optimization_problem_BoP01}) can be transformed into a ``micro" form based on local flows (see \cite{Akamatsu-1997,Bavaud-2012,Saerens-2008}) which will be used later for deriving the sparse RSP. In this new form, one possible way to compute the policy is by sequentially solving the primal and the dual problems. In the case of KL divergence regularization, the algorithm cannot compete against standard procedures developed in \cite{Francoisse-2017,Kivimaki-2012,Saerens-2008}, which are quite efficient. However, the advantage is that this algorithm can be easily adapted to other divergence regularizations.

As shown in \ref{Subsec_alternative_form_RSP01}, the equivalent problem at the local level aims at minimizing the free energy with respect to the transition probabilities instead of paths probabilities,
\begin{equation}
\vline\,\begin{array}{llll}
\minimize\limits_{\mathbf{P}} & \phi^{\mathrm{\textsc{kl}}}_{st}(\mathbf{P}) = \dsum_{(i,j) \in \mathcal{E}}   \bar{n}_{i} p_{ij}  \bigg( c_{ij} + T \log \dfrac{ p_{ij} } { p^{\mathrm{ref}}_{ij}  } \bigg) \\[0.3cm]
\subjectto & \bar{n}_{j} = \dsum_{i \in \mathcal{P}red(j)} \bar{n}_{i} p_{ij} + \delta_{sj}, \text{ for all nodes } j \in \mathcal{V} \\
& \dsum_{j \in \mathcal{S}ucc(i)} p_{ij} = 1 \text{ for all } i \ne t \\
& p_{tj} = 0 \text{ for absorbing and killing node } t \text{ and all } j \in \mathcal{V} \\
& \mathbf{P} \ge \bigzero \\
& \bar{\mathbf{n}} \ge \mathbf{0}
\end{array}
\label{Eq_primal_problem01}
\end{equation}
where $\mathcal{P}red(j)$ is the set of predecessor nodes of node $j$.
This formulation of the optimization problem is derived from Equation (\ref{Eq_local_formulation_objective_function01}), \ref{Subsec_alternative_form_RSP01}, and then solved in \ref{Ap_optimal_policy_computation01}. In this Equation (\ref{Eq_primal_problem01}), the $p_{ij}$ are the elements of the transition matrix $\mathbf{P}$ to be found (the \emph{routing policy}), the $p^{\mathrm{ref}}_{ij}$ are the elements of the transition matrix of the reference random walk on the graph (see Equation (\ref{Eq_Transition_probabilities01})), and the $\bar{n}_{j}$ are the expected numbers of visits to the nodes when walking on $G$ according to the transition matrix $\mathbf{P}$, provided by $\bar{n}_{j} = \sum_{i \in \mathcal{P}red(j)} \bar{n}_{i} p_{ij} + \delta_{sj}$ for nodes in $\mathcal{V}$ (see, e.g., \cite{Norris-1997,Taylor-1996}). Note that because the input flow in source node $s$ is 1 and target node $t$ is the only absorbing and killing node in the network (the sink), $\bar{n}_{t} = 1$.

Following the Lagrange formulation of the problem appearing in \ref{Eq_local_formulation_lagrange_function01} and inspired by discrete space-time optimal control (see, e.g., \cite{Luenberger-2010}), the $p_{ij}$ and the $\bar{n}_{j}$ can be considered as independent (the relations between these variables are encoded in the (linear) constraints). Notice that it is not necessary to impose $p_{ij}=0$ for missing edges as this will be enforced automatically by the algorithm thanks to the use of the KL divergence.

The intuition behind the Equation (\ref{Eq_primal_problem01}) is as follows. The first line (objective function) computes the total expected cost plus KL divergence when following edge $(i,j)$ because $\bar{n}_{i} p_{ij} = \bar{n}_{ij}$ represents the flow in edge $(i,j)$ (expected number of visits to $i$ times the probability of following $(i,j)$). The second line (first constraint) provides the expression for computing the expected number of visits to each node. The last lines state the sum-to-one as well as the non-negativity constraints, which are in fact not necessary in the case of a KL regularization but which will be important in the next section when dealing with Tsallis divergence.

In \ref{Ap_alternative_randomized_shortest_paths01} (see Equations (\ref{Eq_transition_probabilities_computation01}) and (\ref{Eq_free_energy_computation01})), it is shown that the transition probabilities, providing the routing policy, can be computed by sequentially updating the Lagrange parameters $\lambda^{\mathrm{\textsc{kl}}}_{i}$ (associated to the computation of the expected number of visits to nodes) from the transition probabilities and vice-versa thanks to

\begin{itemize}

\item Compute the Lagrange parameters by solving the system of linear equations:
\begin{equation}
\lambda^{\mathrm{\textsc{kl}}}_{i} - \sum_{j \in \mathcal{S}ucc(i)} p_{ij} \lambda^{\mathrm{\textsc{kl}}}_{j}
= \sum_{j \in \mathcal{S}ucc(i)} p_{ij} \bigg( c_{ij} + T \log \frac{ p_{ij} } { p_{ij}^{\mathrm{ref}} } \bigg) \quad \text{for all } i \in \mathcal{V}
\label{Eq_optimal_free_energy_elementwise_computation01}
\end{equation}

\item Compute the transition probabilities:
\begin{equation}
p_{ij} = 
\begin{cases}
 \dfrac{ p_{ij}^{\mathrm{ref}} \exp[-\theta (c_{ij} + \lambda^{\mathrm{\textsc{kl}}}_{j})] } { \dsum_{k \in \mathcal{S}ucc(i)} p_{ik}^{\mathrm{ref}} \exp[-\theta (c_{ik} + \lambda^{\mathrm{\textsc{kl}}}_{k})] } \quad \text{for all } (i,j) \in \mathcal{E} \\
 \qquad 0 \quad \text{for the absorbing and killing target node } i=t
 \end{cases}
 \label{Eq_optimal_transition_probabilities_computation01}
\end{equation}

\end{itemize}
Initially (at iteration 0), the transition probabilities are set to the reference transition probabilities, $p_{ij} = p_{ij}^{\mathrm{ref}}$.

Note that because the target node $t$ is absorbing and killing, Equation (\ref{Eq_optimal_free_energy_elementwise_computation01}) provides $\lambda^{\mathrm{\textsc{kl}}}_{t} = 0$. We also observe that, for missing edges, the transition probabilities are equal to zero, as it should be.
Interestingly, after convergence, the Lagrange parameters are nothing else than the \emph{optimal directed free energy distance}, $\lambda^{\mathrm{\textsc{kl}}}_{i} = \phi^{\mathrm{\textsc{kl}}}_{it}(\mathrm{P}^{*})$, appearing in Equation (\ref{Eq_optimization_problem_BoP01}) and evaluated at the Gibbs-Boltzmann distribution $\mathrm{P}^{*}$ (Equation (\ref{Eq_Boltzmann_probability_distribution01})), as discussed in \ref{Ap_optimal_policy_computation01}.
Moreover, notice that, for an undirected graph and edge costs defined as $c_{ij} = 1/a_{ij}$, the Equation (\ref{Eq_optimal_free_energy_elementwise_computation01}) is nothing more than the harmonic function (induced by Kirchhoff's current law and Ohm's law) computing the voltage associated to the nodes, when considering sources of voltage on nodes \cite{Snell-1984}.

We now follow the same derivation with the difference that we will be using Tsallis instead of KL divergence for defining a sparse policy and the corresponding (directed) FE distance.

\section{Sparse randomized shortest paths}
\label{Sec_sparse_randomized_shortest_paths01}

We now turn to the development of the main contribution of the paper, the introduction of the Tsallis RSP routing framework. Depending on $T$, it provides a kind of ``compressed graph structure" keeping only the most relevant edges for communicating efficiently from source node $s$ to target node $t$ (see Figure \ref{fig:NetFlows} for an illustration).

\subsection{Statement of the problem}

If we have a set of $m$ mutually exclusive random outcomes of a chance experiment whose probabilities $p_{j} \ge 0$ sum to one, as well as known reference probabilities $p_{j}^{\mathrm{ref}} \ge 0$, the Tsallis (\cite{Tsallis-1998,Tsallis-2009} also called Havrda-Charvat \cite{Havrda-1967,Kapur-1989}) directed $r$-divergence between the two probability distributions is given by
\begin{equation}
H_{r}(\mathbf{p} | \mathbf{p}_{\mathrm{ref}})
= \tfrac{1}{r-1} \bigg( \sum_{j=1}^{m} \bigg( \frac{ \hspace{-5pt} p_{j}^{r} } { (p_{j}^{\mathrm{ref}})^{r-1} } \bigg) - 1 \bigg)
= \tfrac{1}{r-1} \sum_{j=1}^{m} p_{j} \bigg( \bigg( \dfrac{ p_{j} } { p^{\mathrm{ref}}_{j} } \bigg)^{\hspace{-3pt} r-1} - 1 \bigg)
\label{Eq_Tsallis_r_divergence01}
\end{equation}
where, in this work, the parameter $r$ will be assumed to be larger than one, $r > 1$, but the most common value is $r=2$. The role of $r$ will be discussed after Equation (\ref{Eq_sparse_optimization_problem_reference01}). This measure  (\ref{Eq_Tsallis_r_divergence01}) generalizes the KL divergence \cite{Kapur-1989} in the sense that it converges to this quantity when $r \rightarrow 1^{+}$. It is also closely related to the Simpson-Gini index widely used in decision trees, as well as other measures of uncertainty and diversity (see, e.g., \cite{Kapur-1989,Keylock-2005}).
As for the more standard KL divergence, this measure is non-negative and quantifies the divergence between the two probability distributions.

The Tsallis divergence will now be used in order to define an optimal, sparse, policy.
From Equation (\ref{Eq_primal_problem01}), the problem aims to minimize the Tsallis-based free energy function instead of the KL-based free energy of Equation (\ref{Eq_optimization_problem_BoP01}),
\begin{equation}
\vline\,\begin{array}{llll}
\minimize\limits_{\mathbf{P}} & \phi^{\mathrm{\textsc{t}s}}_{st}(\mathbf{P}) = \dsum_{(i,j) \in \mathcal{E}}   \bar{n}_{i} p_{ij} \bigg( c_{ij} + \tfrac{T}{r-1} \bigg( \bigg( \dfrac{ p_{ij} } { p^{\mathrm{ref}}_{ij} } \bigg)^{\hspace{-3pt} r-1} - 1 \bigg) \bigg) \\[0.3cm]
\subjectto & \bar{n}_{j} = \dsum_{i \in \mathcal{P}red(j)} \bar{n}_{i} p_{ij} + \delta_{sj}, \text{ for all nodes } j \in \mathcal{V} \\
& \dsum_{j \in \mathcal{S}ucc(i)} p_{ij} = 1 \text{ for all } i \ne t \\
& p_{tj} = 0 \text{ for absorbing and killing node } t \text{ and all } j \in \mathcal{V} \\
& \mathbf{P} \ge \bigzero \\
& \bar{\mathbf{n}} \ge \mathbf{0}
\end{array}
\label{Eq_primal_problem_Tsallis01}
\end{equation}

For convenience, let us renumber the nodes in such a way that node $1$ is the source node and node $n$ (last node) is the target node\footnote{It is assumed that $s \ne t$.}.
The Lagrange function (see \ref{Ap_optimal_policy_computation01}, especially Equation (\ref{Eq_local_formulation_lagrange_function01}), for details) integrating the equality constraints (but not the non-negativity constraints restricting the domain of $\mathbf{P}$ and $\bar{\mathbf{n}}$) becomes
\begin{align}
&\mathscr{L}(\mathbf{P},\bar{\mathbf{n}};\boldsymbol{\mu},\boldsymbol{\lambda}_{\mathrm{\textsc{t}s}})
= \dsum_{i \in \mathcal{V} \setminus n} \bar{n}_{i} \dsum_{j \in \mathcal{S}ucc(i)} p_{ij}  \bigg( c_{ij} + \tfrac{T}{r-1} \bigg( \bigg( \dfrac{ p_{ij} } { p^{\mathrm{ref}}_{ij} } \bigg)^{\hspace{-3pt} r-1} - 1 \bigg) \bigg)  \nonumber \\
& \qquad \qquad + \dsum_{j \in \mathcal{V}} \lambda^{\mathrm{\textsc{t}s}}_{j} \bigg( \dsum_{i \in \mathcal{P}red(j)} \bar{n}_{i} p_{ij} + \delta_{1j} - \bar{n}_{j} \bigg)
+ \dsum_{i \in \mathcal{V} \setminus n} \mu_{i} \bigg( 1 - \dsum_{j \in \mathcal{S}ucc(i)} p_{ij} \bigg)
\label{Eq_Tsallis_formulation_lagrange_function01}
\end{align}
where we also have to deal with non-negativity constraints on the transition probabilities, $p_{ij} \in \mathbb{R}_{+}$, and the expected number of visits, $\bar{n}_{i} \in \mathbb{R}_{+}$. As for the standard RSP, for $\theta = 1/T \rightarrow 0^{+}$, this model becomes a simple absorbing random walk on $G$ with transition probabilities provided by Equation (\ref{Eq_Transition_probabilities01}) (reference probabilities). For $\theta \rightarrow \infty$, it provides least-cost routing concentrated on shortest paths.

Let us proceed like in previous section for optimizing the objective function with respect to the transition probabilities. As before, it is not necessary to impose $p_{ij}=0$ for missing edges because these constraints will be satisfied automatically when recomputing the transition probabilities.

But first, a remark concerning related work before proceeding. If we compare the objective function used in (\ref{Eq_Tsallis_formulation_lagrange_function01}) with the one studied in \cite{Li-2011,Li-2013} (see Equation (18) of \cite{Li-2013} -- also leading to a sparse policy), because the flow in edge $(i,j)$ is provided by $\bar{n}_{ij} = \bar{n}_{i} p_{ij}$, we observe that they are quite different. Indeed, the objective function of \cite{Li-2011,Li-2013} can be rewritten in our notation as
$
c'_{ij} = \sum_{i \in \mathcal{V}} \sum_{j \in \mathcal{S}ucc(i)} \bar{n}_{i} p_{ij} ( c_{ij} + \gamma \, \bar{n}_{i} p_{ij} )
$
where $\gamma \ge 0$ is a parameter controlling sparseness. Note also that \cite{Li-2011,Li-2013} consider a symmetric cost matrix and net flows, instead of raw flows in the present paper.

%
As for previous section, and as detailed in \ref{Ap_alternative_randomized_shortest_paths01}, our optimization procedure optimizes sequentially the objective function by Lagrange duality \cite{Culioli-2012,Griva-2008,Minoux-1986} and a variant of the Arrow-Hurwicz-Uzawa algorithm \cite{Arrow-1958}. More precisely, the Lagrange function is first minimized with respect to the transition probabilities $p_{ij}$ subject to their constraints while considering the Lagrange parameters $\lambda^{\mathrm{\textsc{t}s}}_{i}$ as fixed. Then, Lagrange parameters are computed by maximizing the dual problem. The two steps are iterated until convergence to a stationary point of the objective function (\ref{Eq_primal_problem_Tsallis01}), which is guaranteed because each sub-problem reaches its optimum uniquely \cite{Bertsekas-1999}. A discussion of the convexity of the objective function with respect of the edge flows appears in \ref{Ap_convexity_objective_function01}. In short, it is shown by heuristic arguments that the objective function appearing in Equation (\ref{Eq_primal_problem_Tsallis01}) is convex with respect to the edge flows $\bar{n}_{ij}$, although a formal proof is left for further work. Then, by using the same reasoning as for the KL divergence case of Subsection \ref{Subsec_alternative_form_RSP01}, convergence to a global minimum should be guaranteed.
In practice, for all our experimental runs, we observed that the duality gap is equal to zero, showing that a global minimum is reached.

 Notice that there is an important difference with the previous section (KL regularization), namely that the non-negativity of the transition probabilities is no more guaranteed. We therefore have to deal with these inequality constraints and rely on the Karush-Kuhn-Tucker conditions.

\subsection{Computation of the transition probabilities}

The Lagrange function will first be minimized with respect to the transition probabilities (subject to their constraints), with the Lagrange parameters fixed.
To this end, let us re-arrange the Lagrange function (\ref{Eq_Tsallis_formulation_lagrange_function01})
%
%
by gathering all the terms depending on the transition probabilities,
\begin{align}
&\mathscr{L}(\mathbf{P},\bar{\mathbf{n}};\boldsymbol{\mu},\boldsymbol{\lambda}_{\mathrm{\textsc{t}s}}) \nonumber \\
&= \dsum_{i=1}^{n-1} \bar{n}_{i}  \bigg( \dsum_{j=1}^{n} p_{ij} \underbracket[0.5pt][3pt]{ \big( c_{ij} + \lambda^{\mathrm{\textsc{t}s}}_{j} \big) }_{ \text{augmented costs } c'_{ij}} + \underbracket[0.5pt][3pt]{  \tfrac{T}{r-1} \dsum_{j=1}^{n} p_{ij} \bigg( \dfrac{ p_{ij} } { p^{\mathrm{ref}}_{ij} } \bigg)^{\hspace{-3pt} r-1} }_{ \text{regularization term} } \bigg)
+ \dsum_{i=1}^{n-1} \mu_{i} \bigg( 1 - \dsum_{j=1}^{n} p_{ij} \bigg) \nonumber \\
&\quad  - (\bar{n}_{\bullet} - 1) + \dsum_{i=1}^{n} \dsum_{j=1}^{n} \lambda^{\mathrm{\textsc{t}s}}_{j} ( \delta_{1j} - \bar{n}_{j} )
\label{Eq_Tsallis_formulation_lagrange_function_reorganized01}
\end{align}
because $\bar{n}_{n} = 1$ and $\bar{n}_{\bullet} = \sum_{i=1}^{n} \bar{n}_{i}$. In this last expression, we defined the \emph{augmented costs} for each row $i$ (and thus each transient node $i$), considered as a column vector, $\mathbf{c}'_{i} = \mathbf{row}_{i}(\mathbf{C}) + \boldsymbol{\lambda}_{\mathrm{\textsc{t}s}}$.

From (\ref{Eq_Tsallis_formulation_lagrange_function_reorganized01}), we observe that the part of the objective function depending on the transition probabilities (first line) is a sum of $(n-1)$ terms that can be optimized independently at the level of each node and its incident links. The same is true for the constraints (sum-to-one and non-negativity) which are also operating at the level of the nodes. Moreover, the expected numbers of visits $\bar{n}_{i}$ are considered as independent from the transition probabilities because of the Lagrangian formulation of the problem.

Therefore, the problem that needs to be solved for each transient node $i = 1, \dots, (n-1)$ in turn is
\begin{equation}
\vline\,\begin{array}{llll}
\minimize\limits_{\mathbf{p}_{i}} & \underbracket[0.5pt][3pt]{  (\mathbf{c}_{i}')^{\mathrm{T}} \mathbf{p}_{i}  }_{ \text{expected cost}}  +  \underbracket[0.5pt][3pt]{   \tfrac{ T } { r-1 } \, \mathbf{p}_{i}^{\mathrm{T}} ( \mathbf{p}_{i} \div \mathbf{p}_{i}^{\mathrm{ref}} )^{(r-1)}  }_{ \text{$r$-power regularization term}}    \\[0.6cm]
\subjectto & \mathbf{e}^{\mathrm{T}} \mathbf{p}_{i} = 1 \\
           & \mathbf{p}_{i} \ge \mathbf{0}
\end{array}
\label{Eq_sparse_optimization_problem_reference01}
\end{equation}
together with
\begin{equation}
\begin{cases}
\mathbf{c}'_{i} = \mathbf{row}_{i}(\mathbf{C}) + \boldsymbol{\lambda}_{\mathrm{\textsc{t}s}} \\
\mathbf{p}_{i}^{\mathrm{ref}} = \mathbf{row}_{i}(\mathbf{P}_{\mathrm{ref}}) \\
r > 1
\end{cases} \nonumber
\end{equation}
where $\div$ is the elementwise division, $(r-1)$ is the elementwise power and $T = 1/\theta > 0$ monitors the balance between expected augmented cost and Tsallis divergence minimization. This shows that the Lagrange parameters $\lambda^{\mathrm{\textsc{t}s}}_{j}$ can be interpreted as the virtual costs that have to be added to the real costs $c_{ij}$ in order to be able to compute the local transition probabilities at the level of each node thanks to Equation (\ref{Eq_sparse_optimization_problem_reference01}).

Note that the $r$ parameter controls the impact of the costs on the probabilities. For example, for uniform reference probabilities and linear costs, $\mathbf{c} = [1, 2, 3, 4, 5]^{\mathrm{T}}$, the form of the resulting probability distribution will be decreasing and convex when $r < 2$ ($\mathbf{p} = [0.480,0.295,0.156,0.0602,0.00927]^{\mathrm{T}}$ when $r=1.5$ and $T=1$), linear when $r = 2$ ($\mathbf{p} = [0.4,0.3,0.2,0.1,0.0]^{\mathrm{T}}$ with $T=1$), and concave when $r > 2$ ($\mathbf{p} = [0.289,0.262,0.229,0.182,0.0375]^{\mathrm{T}}$ when $r=4$ and $T=1$).
On the other hand, the temperature $T$ mainly controls the sparseness of the distribution ($\mathbf{p} = [0.533,0.333,0.133,0.0,0.0]^{\mathrm{T}}$ when $r=2$ with $T=0.5$ and $\mathbf{p} = [0.240,0.220,0.200,0.180,0.160]^{\mathrm{T}}$ when $r=2$ with $T=5$).

Thus, the problem of computing the transition probabilities reduces to the minimization of a linear function with a $r$-power regularization term on the probability simplex. Interestingly, for a uniform reference distribution, this problem of minimizing a linear objective function with quadratic as well as $r$-power regularization has been studied in the fuzzy clustering literature \cite{Kanzawa-2013,Miyamoto-1998}. In this context, the objective was to fuzzify the membership functions thanks to a quadratic regularization, similar to the Tsallis entropy, in a fuzzy $k$-means. Note that in their original work, the constant term $-1$ (minus one) present in the Tsallis entropy is not taken into account -- the authors simply consider a quadratic regularization, which has no effect on the solution. Note that \cite{Miyamoto-1998} solves the quadratic regularization problem with $r=2$ while \cite{Kanzawa-2013} provides an algorithm for the more general case $r>1$.

Inspired by this previous work, the problem (\ref{Eq_sparse_optimization_problem_reference01}) is solved by a procedure (spmin) computing a sparse minimum derived in \ref{Appendix_B} for each node $i \ne t$,
\begin{equation}
\mathbf{p}_{i}^{*} = \sparsemin(\mathbf{c}'_{i},\mathbf{p}_{i}^{\mathrm{ref}},r,T) \quad \text{for } i = 1, \dots, (n-1)
\label{Eq_sparse_min_reference01}
\end{equation}
which returns a possibly sparse discrete probability distribution $\mathbf{p}_{i}^{*}$. In short, the solution for the general case is of the form (Equation (\ref{Eq_KKT_result_binary_reference01}))
\begin{equation}
\mathbf{p}^{*}_{i} =
\mathbf{p}_{i}^{\mathrm{ref}} \circ \sqrt[(r-1)]{ \tfrac{r-1}{r T} [ \mu_{i} \mathbf{e} - \mathbf{c}'_{i} ]_{+} }
\label{Eq_final_expression_Tsallis_transition_probabilities01}
\end{equation}
where $[x]_{+} = \max(x,0)$: if $x$ is negative, it is put to $0$, and $\circ$ is the elementwise Hadamard product. The vector $\mathbf{e}$ is a column vector full of $1$'s and the threshold $\mu_{i}$ must be chosen in order to satisfy the sum-to-one constraint, $(\mathbf{p}^{*}_{i})^{\mathrm{T}} \mathbf{e} = 1$.
The expression (\ref{Eq_final_expression_Tsallis_transition_probabilities01}) is the counterpart of Equation (\ref{Eq_path_transition_probabilities01}) based on the KL divergence.

The procedure (\ref{Eq_sparse_min_reference01}) is run on every node $i \ne n$ in order to obtain an updated transition matrix $\mathbf{P}$. Again, the transition probabilities only depend on the Lagrange parameters $\boldsymbol{\lambda}_{\mathrm{\textsc{t}s}}$ through the augmented costs.

Based on previous work \cite{Kanzawa-2013,Miyamoto-1998}, three algorithms are developed in the \ref{Appendix_B}. More precisely, in \ref{Appendix_B1}, we consider the special case of a quadratic regularization term, $r=2$, whereas the general case $r > 1$ is developed in \ref{Appendix_B2}. This is because the quadratic case is simpler and leads to an efficient algorithm based on a linear search once the nodes have been sorted by increasing cost, while the general case is handled by using a bisection search which turns out to be slower in practice. Our contribution with respect to \cite{Kanzawa-2013,Miyamoto-1998} is the introduction of non-uniform reference probabilities in \ref{Appendix_B3} which allow to deal with the Tsallis divergence instead of the Tsallis entropy.
Let us now turn to the computation of the Lagrange parameters.

\subsection{Computation of the Lagrange parameters}

The second step, i.e.\ the computation of the Lagrange parameter vector $\boldsymbol{\lambda}_{\mathrm{\textsc{t}s}}$, follows the same principle as for the KL divergence (see Equation (\ref{Eq_optimal_free_energy_elementwise_computation01}) and Subsection \ref{Ap_optimal_policy_computation01} for details), the only difference being the definition of the divergence. Indeed, the elements of $\tilde{\mathbf{h}}_{\mathrm{\textsc{kl}}}$, based on the KL divergence, in Equation (\ref{Eq_dual_problem01}), now become $\tilde{h}_{i}^{\mathrm{\textsc{t}s}} = \frac{1}{r-1} \sum_{j \in \mathcal{S}ucc(i)} p_{ij} \big( ( p_{ij} / p^{\mathrm{ref}}_{ij} )^{r-1} - 1 \big)$ for the Tsallis divergence.

By similitude with the KL regularization (see Equation (\ref{Eq_optimal_free_energy_elementwise_computation01})), we define the \emph{Tsallis directed free energy} potential as the values of the Lagrange parameters obtained after convergence of the optimization problem,
\begin{equation}
\boldsymbol{\phi}_{\mathrm{\textsc{t}s}} \triangleq \boldsymbol{\lambda}_{\mathrm{\textsc{t}s}}
\label{Eq_Tsallis_directed_free_energy_Lagrange01}
\end{equation}
and we verified that they correspond to the minimized free energy objective function (\ref{Eq_primal_problem_Tsallis01}). These quantities are obtained by solving the following system of linear equations
\begin{equation}
\lambda^{\mathrm{\textsc{t}s}}_{i} - \sum_{j \in \mathcal{S}ucc(i)} p_{ij} \lambda^{\mathrm{\textsc{t}s}}_{j} = \sum_{j \in \mathcal{S}ucc(i)} p_{ij} \bigg[  c_{ij} + \tfrac{T}{r-1} \bigg( \bigg( \dfrac{ p_{ij} } { p^{\mathrm{ref}}_{ij} } \bigg)^{\hspace{-3pt} r-1} - 1 \bigg) \bigg] \quad \text{for all } i \in \mathcal{V}
\label{Eq_Tsallis_free_energy_elementwise_computation02}
\end{equation}
with respect to $\boldsymbol{\lambda}_{\mathrm{\textsc{t}s}}$. Note that these equations imply $\lambda^{\mathrm{\textsc{t}s}}_{t} = 0$ for target node $t$. In matrix form, we have
\begin{equation}
(\mathbf{I} - \mathbf{P}) \boldsymbol{\lambda}_{\mathrm{\textsc{t}s}} = \tilde{\mathbf{c}} + T \tilde{\mathbf{h}}_{\mathrm{\textsc{t}s}}
\label{Eq_Tsallis_free_energy_elementwise_computation01}
\end{equation}
where $\tilde{\mathbf{c}} = (\mathbf{P} \circ \mathbf{C}) \mathbf{e}$ and $\tilde{\mathbf{h}}_{\mathrm{\textsc{t}s}} = \frac{1}{r-1} \big[ \mathbf{P} \circ \big( ( \mathbf{P} \div \mathbf{P}_{\mathrm{ref}})^{(r-1)} - \mathbf{E} \big) \big] \mathbf{e}$, $\mathbf{e}$ is a column vector containing $1$ on each row, $\mathbf{E}$ is a square matrix full of 1's, $\circ$ is the elementwise matrix product, and $(r-1)$ is the elementwise power.

Finally, after initializing the transition probabilities to the reference probabilities, the overall procedure for computing the optimal randomized policy based on Tsallis divergence aims at iterating Equations (\ref{Eq_Tsallis_free_energy_elementwise_computation01}) and (\ref{Eq_sparse_min_reference01}) until convergence. Each of the two steps has a unique optimal solution so that the iterative procedure converges \cite{Bertsekas-1999}. As before, we observed that, in all our runs, the duality gap is always zero, showing that a global minimum is reached.
After convergence, the algorithm provides an optimal, possibly sparse, routing policy taking the form of Markov chain with absorbing, killing, node $n$ (or $t$ before renumbering the nodes).

\section{Derived dissimilarities}
\label{Sec_distances01}

We now define two new dissimilarity measures between nodes interpolating between the least cost and the commute-cost\footnote{And thus also the resistance distance based on the effective resistance when $c_{ij}$ is defined as $1/a_{ij}$, like in electrical networks \cite{FoussKDE-2005}. This property only holds in the case of an undirected graph}, up to a constant scaling factor. These quantities are the counterparts of the RSP dissimilarity and the FE distance based on KL regularization.
As for the KL divergence \cite{Francoisse-2017,Kivimaki-2012,Yen-08K}, the quantity $\phi_{it}^{\mathrm{\textsc{t}s}}$, provided by the Lagrange parameter $\lambda_{i}^{\mathrm{\textsc{t}s}}$ after convergence (see Equation (\ref{Eq_Tsallis_directed_free_energy_Lagrange01})), is called the Tsallis directed FE from node $i$ to absorbing, killing, node\footnote{Recall that the target node is always transformed into an absorbing and killing node.} $t$. Accordingly, the \emph{Tsallis FE distance} is the symmetrized quantity
\begin{equation}
\myDelta^{\mathrm{\textsc{t}s\textsc{fe}}}_{st} = \dfrac{\phi_{st}^{\mathrm{\textsc{t}s}} + \phi_{ts}^{\mathrm{\textsc{t}s}}}{2}
\end{equation}
Interestingly, we found that this quantity satisfies the triangle inequality on all the investigated datasets and values of the $\theta$ parameter (see the next, experimental, section). We therefore conjecture that it defines a distance measure between nodes (as in the case for the KL divergence \cite{Francoisse-2017}) and hope to prove it in future work.  

Moreover, the directed Tsallis RSP dissimilarity is based on the total expected cost for reaching target node $t$ from node $s$ when following the optimal policy, which is given by
\begin{equation}
\langle c \rangle_{st} = \dsum_{(i,j) \in \mathcal{E}} \bar{n}_{i} p_{ij} c_{ij}
= \dsum_{(i,j) \in \mathcal{E}} \bar{n}_{ij} c_{ij}
\end{equation}
where $p_{ij}$ is computed by Equation (\ref{Eq_sparse_min_reference01}). The \emph{Tsallis randomized shortest paths dissimilarity} is directly deduced from the previous expression,
\begin{equation}
\myDelta^{\mathrm{\textsc{t}s\textsc{rsp}}}_{st} = \langle c \rangle_{st} + \langle c \rangle_{ts}
\end{equation}
This dissimilarity, however, does not satisfy the triangle inequality for all values of the parameter $\theta$; we are thus in the same situation as for the KL regularization: the Tsallis randomized shortest paths dissimilarity is not a distance measure.
The FE distance and the RSP dissimilarity will be compared on pattern recognition tasks in the next section.

\section{Experiments}
\label{Sec_experiments01}

In this section, we first present an illustrative example of the sparsity of the routing policy induced by the Tsallis-based RSP, by visualizing the net flows on a weighted undirected graph. For each edge $(i,j)$, the \emph{net flow} is defined as $net_{ij} = \max(\bar{n}_{ij} - \bar{n}_{ji},0)$ and is oriented in the direction of the positive flow (at most one per edge). Then, we evaluate our methods by comparing them with other state-of-the-art dissimilarity measures on node clustering and semi-supervised classification tasks\footnote{Note that all results were obtained with Matlab (version R2019) running on an Intel Xeon with $2 \times 8$ core processors of 3.6 GHz and 128 GB of RAM.}.
We have to stress that our goal here is not to propose new node clustering or semi-supervised classification algorithms outperforming state-of-the-art techniques. Rather, the aim is to investigate if the Tsallis-based RSP model is able to capture the community structure of networks in an accurate way, compared to other, existing, state-of-the-art dissimilarity measures between nodes.

\subsection{Illustrative example}

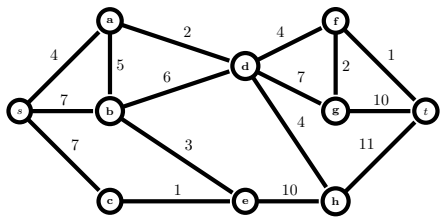
\begin{figure}[t]
\centering
\begin{tikzpicture}[shorten >=1pt, auto, node distance=1cm, ultra thick,
   classic/.style={circle,draw=black,font=\scriptsize\bfseries},
   edge_style/.style={draw=black},scale=0.8,transform shape]
   
    \node[classic] (S) at (1,0) {$s$};
    \node[classic] (a) at (3,2) {a};
    \node[classic] (b) at (3,0) {b};
    \node[classic] (c) at (3,-2) {c};
    \node[classic] (d) at (6,1) {d};
    \node[classic] (e) at (6,-2) {e};
    \node[classic] (f) at (8,2) {f};
    \node[classic] (g) at (8,0) {g};
    \node[classic] (h) at (8,-2) {h};
    \node[classic] (T) at (10,0) {$t$};
    
    
    \draw[edge_style]  (S) edge node{4} (a);
    \draw[edge_style]  (S) edge node{7} (b);
    \draw[edge_style]  (S) edge node{7} (c);
    \draw[edge_style]  (a) edge node{5} (b);
    \draw[edge_style]  (a) edge node{2} (d);
    \draw[edge_style]  (b) edge node{6} (d);
    \draw[edge_style]  (b) edge node{3} (e);
    \draw[edge_style]  (c) edge node{1} (e);
    \draw[edge_style]  (d) edge node{4} (f);
    \draw[edge_style]  (d) edge node{7} (g);
    \draw[edge_style]  (d) edge node{4} (h);
    \draw[edge_style]  (e) edge node{10} (h);
    \draw[edge_style]  (f) edge node{2} (g);
    \draw[edge_style]  (f) edge node{1} (T);
    \draw[edge_style]  (g) edge node{10} (T);
    \draw[edge_style]  (h) edge node{11} (T);
    
 
\end{tikzpicture}
\caption{A small undirected graph composed of 10 nodes with costs on edges; adapted from \cite{Price-1971}.}
\label{fig:GraphIllustr}
\end{figure}

Consider the graph represented in Figure \ref{fig:GraphIllustr}, containing a source node $s$ and a target node $t$. The weight on the edges represents the cost associated to the transitions.

\begin{figure}[t]
\subfigure[$\theta=0.2$]{
\resizebox{.49\linewidth}{!}{
        \scriptsize
\centering
\begin{tikzpicture}[shorten >=1pt, auto, node distance=1cm, thick,
   classic/.style={circle,draw=black,font=\scriptsize\bfseries},
   edge_style/.style={draw=black},scale=0.8,transform shape]
   
   \node[classic] (S) at (1,0) {$s$};
    \node[classic] (a) at (3,2) {a};
    \node[classic] (b) at (3,0) {b};
    \node[classic] (c) at (3,-2) {c};
    \node[classic] (d) at (6,1) {d};
    \node[classic] (e) at (6,-2) {e};
    \node[classic] (f) at (8,2) {f};
    \node[classic] (g) at (8,0) {g};
    \node[classic] (h) at (8,-2) {h};
    \node[classic] (T) at (10,0) {$t$};
    
    
    \draw[edge_style,->]  (S) edge node[sloped]{0.529} (a);
    \draw[edge_style,->]  (S) edge node{0.348} (b);
    \draw[edge_style,->]  (S) edge node[sloped]{0.123} (c);
    \draw[edge_style,->]  (b) edge node{0.003} (a);
    \draw[edge_style,->]  (a) edge node[sloped]{0.532} (d);
    \draw[edge_style,->]  (b) edge node[sloped]{0.290} (d);
    \draw[edge_style,->]  (b) edge node[sloped]{0.055} (e);
    \draw[edge_style,->]  (c) edge node{0.123} (e);
    \draw[edge_style,->]  (d) edge node[sloped]{0.418} (f);
    \draw[edge_style,->]  (d) edge node[sloped]{0.284} (g);
    \draw[edge_style,->]  (d) edge node[sloped]{0.120} (h);
    \draw[edge_style,->]  (e) edge node{0.178} (h);
    \draw[edge_style,->]  (g) edge node{0.050} (f);
    \draw[edge_style,->]  (f) edge node[sloped]{0.468} (T);
    \draw[edge_style,->]  (g) edge node{0.234} (T);
    \draw[edge_style,->]  (h) edge node[sloped]{0.298} (T);
 
\end{tikzpicture}
    }}
    \hfill
    \subfigure[$\theta=0.5$]{
\resizebox{.49\linewidth}{!}{
        \scriptsize
\centering
\begin{tikzpicture}[shorten >=1pt, auto, node distance=1cm, thick,
   classic/.style={circle,draw=black,font=\scriptsize\bfseries},
   unreached/.style={circle,draw=gray!40,font=\scriptsize\bfseries},
   edge_style/.style={draw=black},scale=0.8,transform shape]
   
   \node[classic] (S) at (1,0) {$s$};
    \node[classic] (a) at (3,2) {a};
    \node[classic] (b) at (3,0) {b};
    \node[unreached] (c) at (3,-2) {c};
    \node[classic] (d) at (6,1) {d};
    \node[classic] (e) at (6,-2) {e};
    \node[classic] (f) at (8,2) {f};
    \node[classic] (g) at (8,0) {g};
    \node[classic] (h) at (8,-2) {h};
    \node[classic] (T) at (10,0) {$t$};
    
    
    \draw[edge_style,->]  (S) edge node[sloped]{0.795} (a);
    \draw[edge_style,->]  (S) edge node{0.205} (b);
    \draw[edge_style,->]  (b) edge node{0.066} (a);
    \draw[edge_style,->]  (a) edge node[sloped]{0.861} (d);
    \draw[edge_style,->]  (b) edge node[sloped]{0.132} (d);
    \draw[edge_style,->]  (b) edge node[sloped]{0.007} (e);
    \draw[edge_style,->]  (d) edge node[sloped]{0.628} (f);
    \draw[edge_style,->]  (d) edge node[sloped]{0.255} (g);
    \draw[edge_style,->]  (d) edge node[sloped]{0.110} (h);
    \draw[edge_style,->]  (e) edge node{0.007} (h);
    \draw[edge_style,->]  (g) edge node{0.138} (f);
    \draw[edge_style,->]  (f) edge node[sloped]{0.766} (T);
    \draw[edge_style,->]  (g) edge node{0.117} (T);
    \draw[edge_style,->]  (h) edge node[sloped]{0.117} (T);
 
\end{tikzpicture}
    }}
    \subfigure[$\theta=1$]{
\resizebox{.49\linewidth}{!}{
     
        \scriptsize
\centering
\begin{tikzpicture}[shorten >=1pt, auto, node distance=1cm, thick,
   classic/.style={circle,draw=black,font=\scriptsize\bfseries},
   unreached/.style={circle,draw=gray!40,font=\scriptsize\bfseries},
   edge_style/.style={draw=black},scale=0.8,transform shape]
   
   \node[classic] (S) at (1,0) {$s$};
    \node[classic] (a) at (3,2) {a};
    \node[unreached] (b) at (3,0) {b};
    \node[unreached] (c) at (3,-2) {c};
    \node[classic] (d) at (6,1) {d};
    \node[unreached] (e) at (6,-2) {e};
    \node[classic] (f) at (8,2) {f};
    \node[classic] (g) at (8,0) {g};
    \node[unreached] (h) at (8,-2) {h};
    \node[classic] (T) at (10,0) {$t$};
    
    
    \draw[edge_style,->]  (S) edge node{1} (a);
    \draw[edge_style,->]  (a) edge node{1} (d);
    \draw[edge_style,->]  (d) edge node[sloped]{0.848} (f);
    \draw[edge_style,->]  (d) edge node[sloped]{0.152} (g);
    \draw[edge_style,->]  (g) edge node{0.139} (f);
    \draw[edge_style,->]  (f) edge node[sloped]{0.987} (T);
    \draw[edge_style,->]  (g) edge node{0.013} (T);
 
\end{tikzpicture}
    
    }}
    \hfill
    \subfigure[$\theta=2$]{
\resizebox{.49\linewidth}{!}{
    
        \scriptsize
\centering
\begin{tikzpicture}[shorten >=1pt, auto, node distance=1cm, thick,
   classic/.style={circle,draw=black,font=\scriptsize\bfseries},
   unreached/.style={circle,draw=gray!40,font=\scriptsize\bfseries},
   edge_style/.style={draw=black},scale=0.8,transform shape]
   
   \node[classic] (S) at (1,0) {$s$};
    \node[classic] (a) at (3,2) {a};
    \node[unreached] (b) at (3,0) {b};
    \node[unreached] (c) at (3,-2) {c};
    \node[classic] (d) at (6,1) {d};
    \node[unreached] (e) at (6,-2) {e};
    \node[classic] (f) at (8,2) {f};
    \node[unreached] (g) at (8,0) {g};
    \node[unreached] (h) at (8,-2) {h};
    \node[classic] (T) at (10,0) {$t$};
    
    
    \draw[edge_style,->]  (S) edge node{1} (a);
    \draw[edge_style,->]  (a) edge node{1} (d);
    \draw[edge_style,->]  (d) edge node{1} (f);
    \draw[edge_style,->]  (f) edge node{1} (T);
 
\end{tikzpicture}
    
    }}
    \caption{Representation of the net flow from $s$ to $t$ in function of $\theta$ for the Tsallis RSP and using $r=2$.}
    \label{fig:NetFlows}
\end{figure}
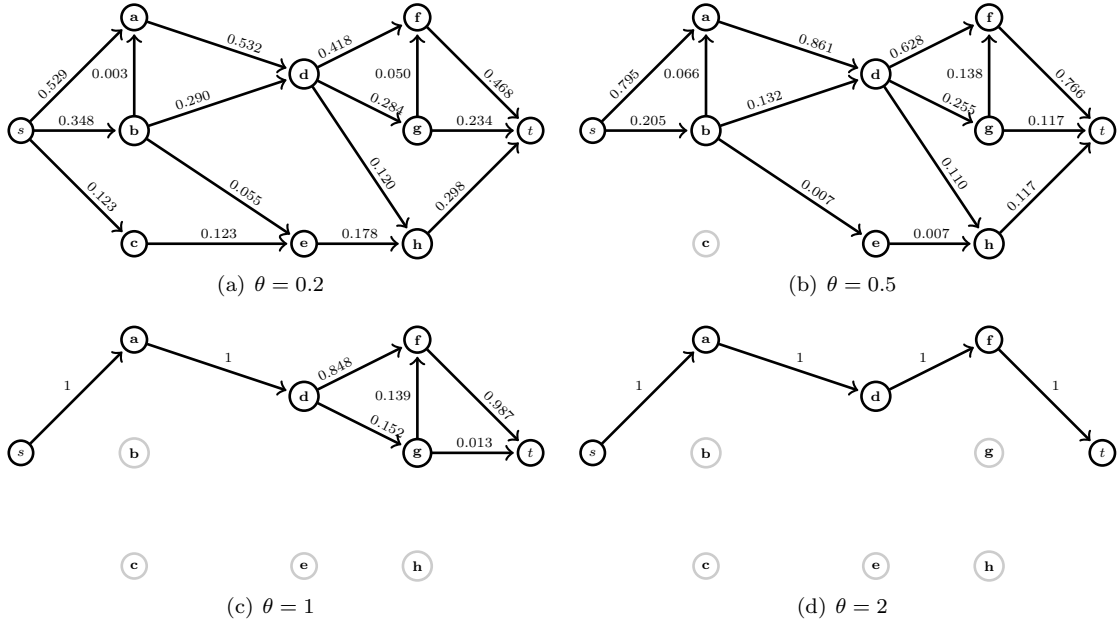

We illustrate how the parameter $\theta$ influences the sparsity of the transition probabilities (the randomized routing policy) by representing the net flow from the source to the target. These flows are depicted for the value $r=2$ in Figure \ref{fig:NetFlows}, where only positive net flows are drawn.

As the value of $\theta$ increases, some edges gradually become unused (the flow in the edge is equal to zero). Eventually, the flow is entirely concentrated on the shortest path (already when $\theta = 2$ in our example). This clearly shows that the RSP routing policy becomes gradually sparser when the parameter $\theta$ increases. This property is also observed for other values of the parameter $r$.

\subsection{Global experimental setup}
The Tsallis FE (FETsallis) and the Tsallis RSP (RSPTsallis) dissimilarities will be assessed in two different contexts: first, a node clustering task and, second, a graph-based semi-supervised classification task aiming to categorize unlabeled nodes. For all RSP-based methods, the reference transition probabilities are set to $p_{ij}^{\mathrm{ref}} = a_{ij} / {\sum_{j'=1}^{n}} a_{ij'}$, corresponding to a natural random walk on the graph (see Equation (\ref{Eq_Transition_probabilities01})). In addition, the costs on the edges are defined as $c_{ij} = 1/a_{ij}$, as for electrical networks.
As part of the experiments, four dissimilarity matrices between nodes as well as four kernels on a graph will be used as baseline methods to assess our methods.

\subsubsection{Baseline dissimilarities between nodes}
\begin{itemize}
    \item The \textbf{Free Energy} distance (FE, based on KL divergence) and the \textbf{Randomized Shortest Paths Dissimilarity} (RSP, also based on KL divergence) depending on an inverse temperature parameter $\theta = 1/T$. Already presented earlier in Section \ref{Sec_standard_randomized_shortest_paths01}, these methods have been shown to perform well in node clustering \cite{Sommer-2016} as well as in semi-supervised classification tasks \cite{Francoisse-2017}.
    \item The \textbf{Logarithmic Forest} distance (LF). Introduced in \cite{Chebotarev-2011}, it relies on the matrix-forest theorem \cite{Chebotarev-1997} and defines a family of distances interpolating (up to a scaling factor) between the shortest-path distance and the resistance distance, depending on a parameter $\alpha$.
    \item The \textbf{Shortest Path} distance (SP). The distance corresponds to the cost along the shortest path between two nodes $i$ and $j$, derived from the cost matrix $\mathbf C$.
\end{itemize}
These dissimilarity matrices are transformed into inner products (kernel matrices) by classical multidimensional scaling (see later).

\subsubsection{Baseline kernels on a graph}
\begin{itemize}
    \item The \textbf{Neumann} kernel \cite{Scholkopf-2002}  (Katz), initially proposed in \cite{Katz-1953} as a method of computing similarities, and defined as $\mathbf K=(\mathbf I-\alpha\mathbf A)^{-1}-\mathbf I$. The $\alpha$ parameter has to be chosen positive and smaller than the inverse of the spectral radium of $\mathbf A$, $\rho(\mathbf A)=\max_i(|\lambda_i|)$.
    \item The \textbf{Logarithmic Communicability} kernel (lCom) proposed in \cite{ivashkin-2016} as the logarithmic version of the exponential diffusion kernel \cite{Kondor-2002}, also known as the communicability measure \cite{Estrada-2008}, $\mathbf K=\ln(\expm{(t\mathbf A)}),\,t>0$, where $\expm$ is the matrix exponential.
    \item The \textbf{Sigmoid Commute Time} kernel. Proposed in \cite{Yen-2007}, it is obtained by applying a sigmoid transform \cite{Scholkopf-2002} on the commute time kernel \cite{FoussKDE-2005}.  A parameter $\alpha$ controls the sharpness of the sigmoid.
\end{itemize}

In addition, the \textbf{Modularity matrix} $\mathbf Q$ is used as last baseline (Modularity). The matrix is computed by $\mathbf Q = \mathbf A-\frac{\mathbf{dd}^{\text T}}{\mathrm{vol}}$ where $\mathbf{d}$ contains the node degrees and the constant $\mathrm{vol}$ is the volume of the graph (see, e.g., \cite{Newman-2018}).

\subsubsection{Datasets}

A collection of 22 datasets, representing labeled networks, is investigated for the experimental comparisons of the dissimilarity measures. The collection includes Zachary's karate club \cite{Zachary1977}, the Dolphin datasets \cite{Lusseau-2003-emergent,Lusseau-2003-bottlenose}, the Football dataset \cite{Newman2002}, the Political books\footnote{Collected by V. Krebs and labelled by M.E. Newman, this dataset is not published, but available for download at \url{http://www-personal.umich.edu/~mejn/netdata/}.}, three LFR benchmarks \cite{lancichinetti-2008}, the WebKB datasets \cite{Macskassy-07}, the IMDB dataset \cite{Macskassy-07}, and 9 Newsgroup datasets \cite{Lang-1995, Yen-2009}.

The list of datasets along with their main characteristics is available in Table \ref{tab:datasets}. Please note that all the datasets have not been used in both clustering and classification context. The two last columns indicate the investigated task for each dataset.

\begin{table}[t]
\centering
\footnotesize
\begin{tabular}{l|l|l|l||c|c}
\hline
\multicolumn{4}{c||}{\textbf{Dataset (labeled network)}}&\multicolumn{2}{c}{\textbf{Task}}\\ \hline
Name         & Labels & Nodes & Edges & Clustering & Classification \\ \hline
Dolphin\_2      & 2        & 62    & 159 & X &    \\ \hline
Dolphin\_4      & 4        & 62    & 159 & X &    \\ \hline
Football        & 12       & 115   & 613 & X &    \\ \hline
LFR1            & 3        & 600   & 6142 & X &   \\ \hline
LFR2            & 6        & 600   & 4807 & X &   \\ \hline
LFR3            & 6        & 600   & 5233 & X &  \\ \hline
IMDB              & 2        & 1126  & 20282 &  & X \\ \hline                                   
Newsgroup\_2\_1 & 2        & 400   & 33854 & X & X \\ \hline
Newsgroup\_2\_2 & 2        & 398   & 21480 & X & X \\ \hline
Newsgroup\_2\_3 & 2        & 399   & 36527 & X & X \\ \hline
Newsgroup\_3\_1 & 3        & 600   & 70591 & X & X \\ \hline
Newsgroup\_3\_2 & 3        & 598   & 68201 & X & X \\ \hline
Newsgroup\_3\_3 & 3        & 595   & 64169 & X & X \\ \hline
Newsgroup\_5\_1 & 5        & 998   & 176962 & X & X \\ \hline
Newsgroup\_5\_2 & 5        & 999   & 164452 & X & X\\ \hline
Newsgroup\_5\_3 & 5        & 997   & 155618 & X & X\\ \hline
Political books & 3        & 105   & 441    & X & \\ \hline
WebKB-Cornell  & 6        & 346   & 13416 &  & X \\ \hline
WebKB-Texas  & 6        & 334   & 16494 &  &  X\\ \hline
WebKB-Washington & 6        & 434   & 15231 &  & X \\ \hline
WebKB-Wisconsin & 6        & 348   & 16625 &  & X \\ \hline
Zachary         & 2        & 34    & 78     & X & \\ \hline
\end{tabular}
\caption{Datasets (networks) used for the experiments.}
\label{tab:datasets}
\end{table}

\subsection{Node clustering experiment}

We first describe the node clustering application together with the experimental methodology.

\subsubsection{Evaluation metrics}

Each partition will be assessed by comparing it with the observed ``true partition" of the dataset. The following standard criteria will be used to evaluate the similarity between both partitions.
\begin{itemize}
    \item The \textbf{Normalized Mutual Information (NMI)} \cite{Fred-2003,Strehl-2002} between two partitions $\mathcal{U}$ and $\mathcal{V}$ is computed by dividing the mutual information \cite{Cover-2006} between the two partitions by the average of the respective entropy of $\mathcal{U}$ and $\mathcal{V}$. See also \cite{Manning-2008}.
    
    \item The \textbf{Adjusted Rand Index (ARI)} \cite{Hubert-1985} is an extension of the Rand Index \cite{Rand-1971}, which measures the degree of overlap between two partitions. The ARI has an expected value of 0, which is not the case for the initial Rand Index.
\end{itemize}

\subsubsection{Experimental methodology}
The experiment relies on a kernel $k$-means (see e.g.,\ \cite{Fouss-2016,Yen-2009}). For the dissimilarities, the followed methodology is similar to the one used in \cite{Sommer-2016} (see this work for more details). More specifically, for each given dataset, the dissimilarity matrix $\mathbf D$ obtained by the different methods is transformed into a kernel $\mathbf K$ (a inner product matrix) using classical multidimensional scaling \cite{Borg-1997}. If the resulting kernel is not positive semi-definite, we simply set the negative eigenvalues to zero when computing the kernel.
As a second step, a kernel $k$-means (see e.g.,\ \cite{Fouss-2016,Yen-2009}) is run 30 times on $\mathbf K$. The NMI and ARI are computed on the partition maximizing the modularity among these 30 trials. Recall that modularity is an unsupervised measure of the quality of a partition of the nodes (a set of communities) \cite{Newman-2018}.

This operation is repeated 30 times (leading to a total of 900 runs of the $k$-means) to obtain the average modularity, NMI and ARI scores over these 30 repetitions for a given method (dissimilarity matrix), with a given value of its parameter (for instance, $\theta$ in the case of methods based on RSP), on a specific dataset. Finally, the reported NMI and ARI score for each method and dataset is the average (over the 30 repetitions) for the parameter value leading to the largest modularity.
Thus, modularity (which is unsupervised) is used as a metrics to tune the parameters of the algorithms  \cite{Sommer-2017}. The parameters that are tuned are the $\theta$ for the FE and the RSP, in both standard (FE and RSP) and Tsallis versions (FETsallis and RSPTsallis), the $\alpha$ for the LF, the $\alpha$ for Katz, the $t$ for the lCom, and finally the $\alpha$ for the sigmoid transform of the SCT. The range of values that are tested strongly differs from the standard to the Tsallis version as we observed that the Tsallis version is less sensitive to variations in $\theta$. The values tested for these parameter are listed in the Table \ref{tab:ParamTuning}.

\begin{table}[t]
\footnotesize
\centering
\begin{tabular}{|l|l|}
\hline
Algorithm                                               & Parameter values \\ \hline
\begin{tabular}[c]{@{}l@{}}FE\\ RSP\end{tabular} & $\theta= (0.001,\,0.005,\,0.01,\,0.05,\,0.1,\,0.5,\,1,\,3,\,5,\,10,\,15,\,20)$\\ \hline
\begin{tabular}[c]{@{}l@{}}FETsallis\\ RSPTsallis\end{tabular} & $\theta=($ $10^{-4}$, $10^{-3}$, $10^{-2}$, $10^{-1}$, $1$, $10$, $10^2$, $10^3$, $10^4$, $10^5)$\\ \hline
LF                                                      & $\alpha= (0.001,\,0.005,\,0.01,\,0.05,\,0.1,\,0.5,\,1,\,3,\,5,\,10,\,15,\,20)$\\ \hline
Katz                                                    & $\alpha=(0.05,\,0.10,\dots,\,0.95) \times (\rho(\mathbf A))^{-1}$\\ \hline
lCom                                                    & $t=(0.01,\,0.02,\,0.05,\,0.1,\,0.2,\,0.5,\,1,\,2,\,5,\,10)$\\ \hline
\begin{tabular}[c]{@{}l@{}}SCT
\end{tabular}      & $\alpha=(5,\,10,\,15,\,\dots,\,50)$\\ \hline
\end{tabular}
\caption{Parameter range for the investigated methods.}
\label{tab:ParamTuning}
\end{table}
For the parameter $r$ from the Tsallis regularization, three different values are tested $r=\{1.5, 2, 3\}$. Only a few values are investigated because the computation of the Tsallis-based dissimilarities is much slower that the one based on the KL divergence.
Moreover, the results for these three values are displayed separately in order to analyse their impact on the results.

\subsubsection{Experimental results}

\begin{figure}[t]
    \centering
    \subfigure[]{\includegraphics*[width=0.49\textwidth,trim= 0 20 0 0]{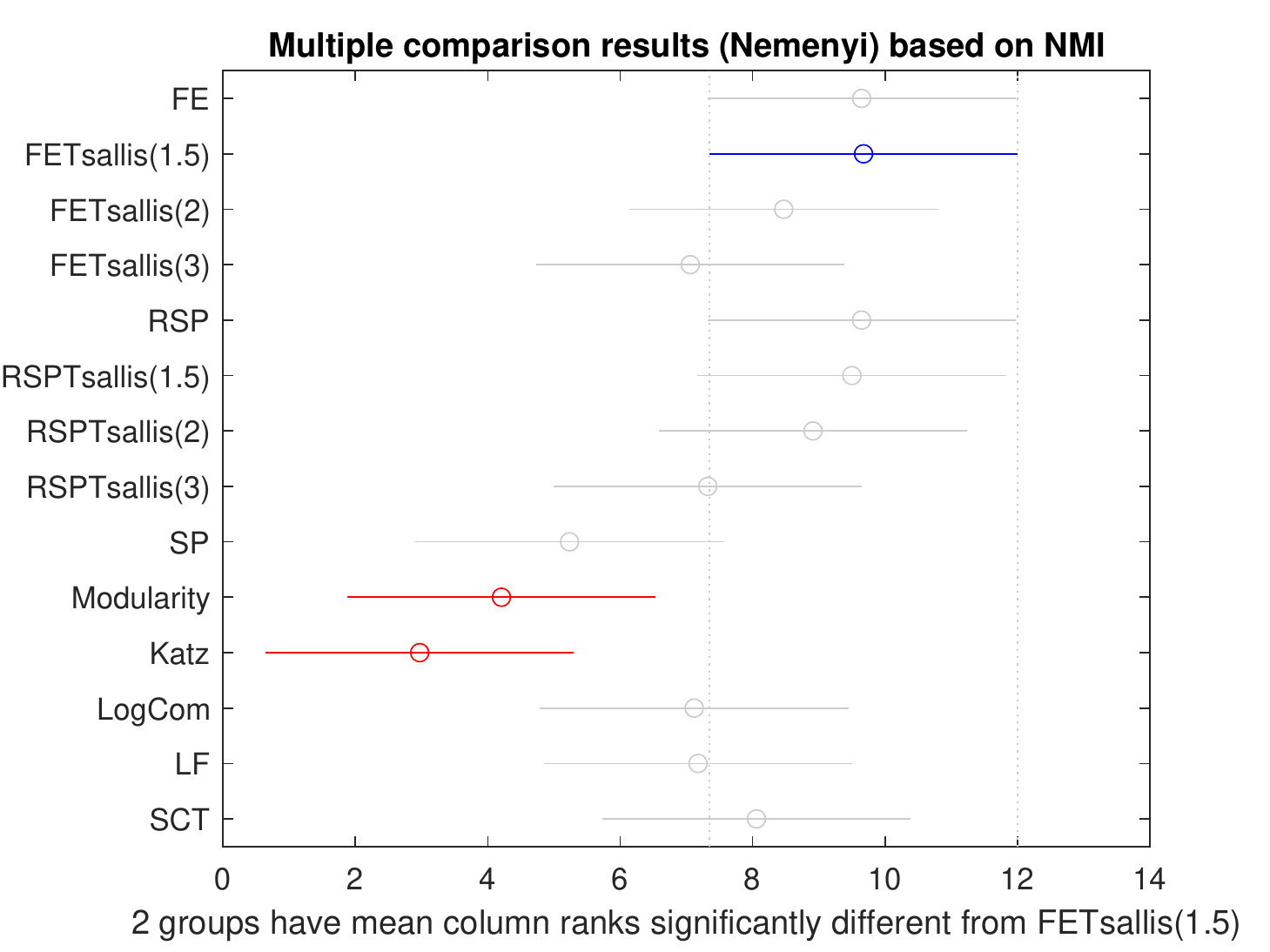}}
    \subfigure[]{
    \includegraphics*[width=0.49\textwidth,trim= 0 20 0 0]{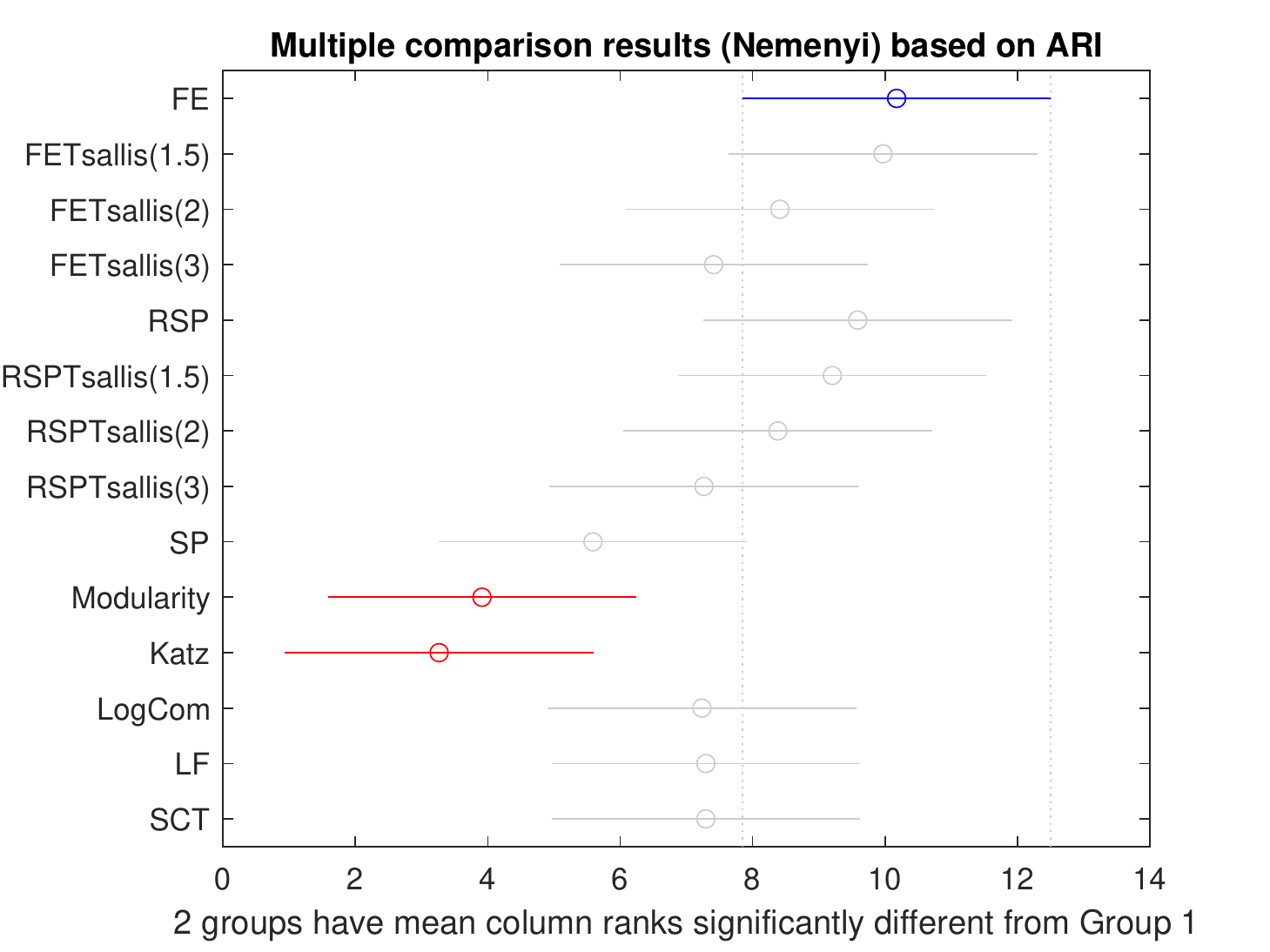}}
    \caption{\footnotesize{Clustering experiment. Mean ranks and 95\% Nemenyi confidence intervals for the 11 methods across the 17 datasets, according to the NMI (a) and the ARI (b) performance measures. Two methods are considered as significantly different if their confidence intervals do not overlap. The best method is highlighted.}}
    \label{fig:nemenyiclustering}
\end{figure}

The different methods are assessed globally across all datasets using the same method as in \cite{Sommer-2016}, based on a non-parametric Friedman-Nemenyi test \cite{Demvsar-2006}. The results for the ARI and the NMI are shown in Figure \ref{fig:nemenyiclustering}. In addition, a Wilcoxon signed-rank test \cite{Wilcoxon-1945} is performed pairwise to measure the significance (at level $\alpha = 0.05$) of the differences observed in the algorithms' performance.

The figure shows that using a parameter $r=1.5$ for the Tsallis regularization tends to yield the best results out of the three values of $r$ that were investigated.
What concerns the FE distance, this $r$ value allows to slightly outperform the FE relying on KL divergence based on the NMI, but it it not the case when considering the ARI. In both cases, the difference is not significant according to the Wilcoxon signed-rank test.
More generally, except for the value $r=3$, no significant difference can be observed, in terms of ranks, between the KL and the Tsallis divergence regularization for both the FE and the RSP from the pairwise Wilcoxon signed-rank tests.

The FETsallis with $r=1.5$ significantly outperforms the Modularity and the Katz kernel according to the Friedman-Nemenyi test. Additionally, according to the Wilcoxon, the difference in performance with the SP, the LogCom and the LF are significant as well.

Thus, what concerns the clustering task and the investigated datasets, the Tsallis regularization yields competitive results with respect to methods that have been shown to perform well in a context of kernel $k$-means clustering \cite{Sommer-2016}.

\subsection{Semi-supervised classification experiment}

We now turn to the semi-supervised classification experiment.

\subsubsection{Evaluation metrics}

Each method will be evaluated in terms of classification accuracy on semi-supervised tasks where a subset of nodes of the graph is kept unlabeled (i.e.\ hidden). Then, the predicted labels of these unlabeled nodes are compared to the true, observed, labels which were hidden.

\subsubsection{Experimental methodology}

We followed the same experimental methodology as in \cite{Francoisse-2017,Guex-2019Cov}. More precisely, this graph-based semi-supervised classification methodology consists in extracting the five\footnote{We arbitrarily report the results for 5 dimensions but also performed experiments with more dimensions with similar conclusions.} dominant eigenvectors of a kernel matrix, derived from the dissimilarity matrix by classical multidimensional scaling, in order to use them as node features in a linear support vector machine (SVM). Note that this setting is inspired by the work of Zhang et al.\ \cite{Zhang-2008b,Zhang-2008} as well as Tang et al.\ \cite{Tang-2009,Tang-2009b,Tang-2010} who compute the dominant eigenvectors (a ``latent social space") of graph kernels or similarity matrices and then input them into a supervised classification method, such as a logistic regression or a SVM, to categorize the nodes.

All the methods are tested by using a standard $5 \times 5$ nested cross-validation methodology. Each external cross-validation contains 5 folds, and methods are tested with a labelling rate of $20\%$. To tune parameters (see Table \ref{tab:ParamTuning} for values), an internal 5-fold cross-validation on the training fold is performed with a labelling rate of $80\%$. The whole cross-validation procedure is repeated 5 times for different random permutations of the data, inducing different sets of labeled/unlabeled nodes. The \emph{final accuracy} of the classifier on the investigated dataset is then obtained by averaging the results over the five repetitions, and is reported in Table \ref{Tab:Acc}. 

Concerning the parameter $r$ of the Tsallis regularization, as for clustering, we tested three values $r = \{1.5, 2, 3\}$. For the SVM, the margin parameter is tuned on the set of values $c = \{10^{-2},10^{-1}, 1, 10, 100\}$. 

\subsubsection{Experimental results}

\begin{figure}[t]
    \centering
    \includegraphics*[width=0.6\textwidth,trim= 0 20 0 0]{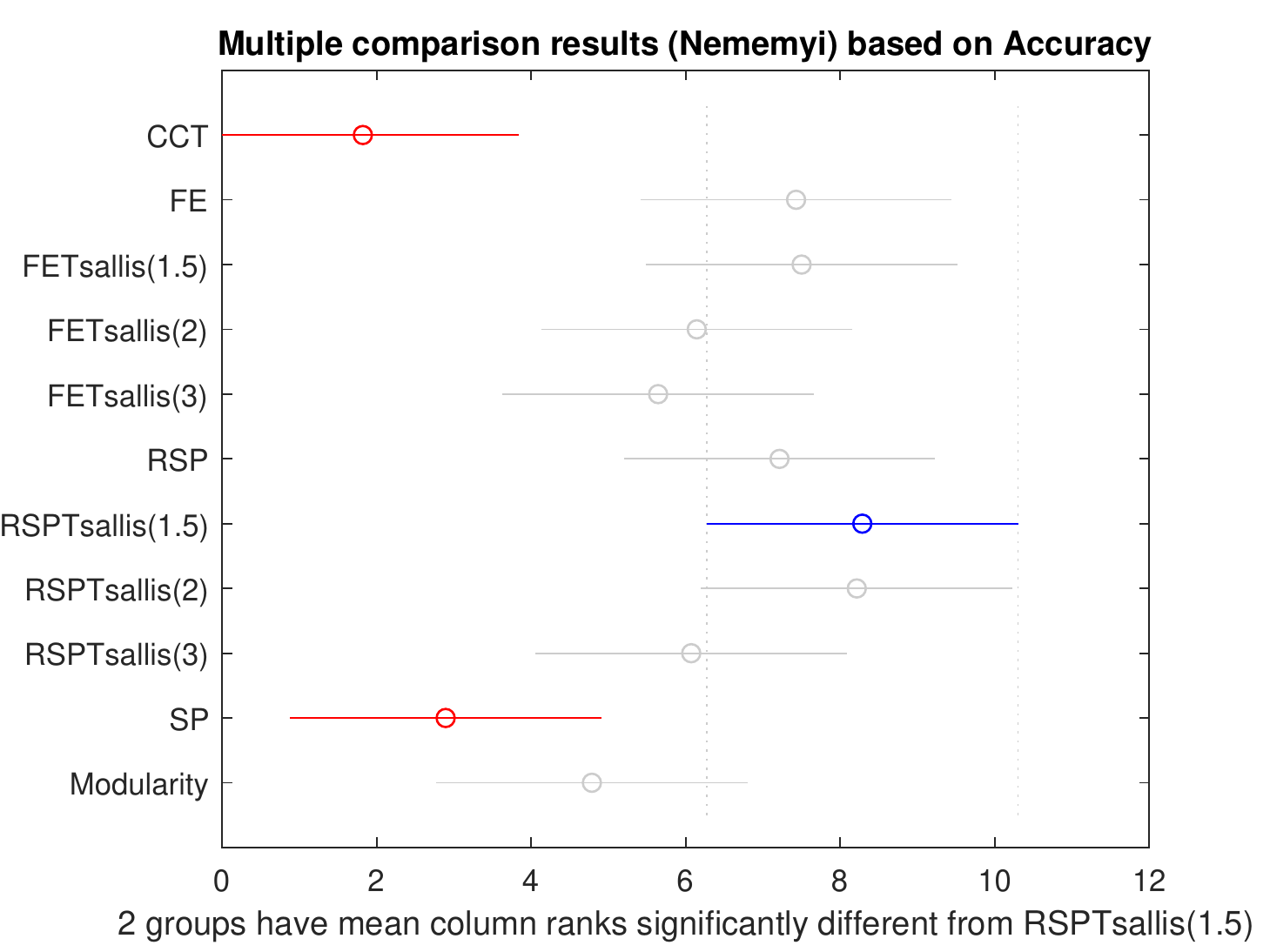}
    \caption{\footnotesize{Semi-supervised experiment. Mean ranks and 95\% Nemenyi confidence intervals for the 14 methods (see Table \ref{Tab:Acc}) across the 14 datasets. Two methods are considered as significantly different if their confidence intervals do not overlap. The best method overall (RSPTsallis(1.5)) is highlighted.}}
    \label{fig:nemenyiacc}
\end{figure}

The results of this experiment are reported in Table \ref{Tab:Acc}. The highest accuracy is highlighted in boldface for each dataset. As can be seen, the best method is dataset-dependent and no obvious, global, pattern is present. 
Therefore, in order to rate globally the results of each method, as before, we perform a nonparametric Friedman-Nemenyi statistical test and a Wilcoxon signed-rank tests at a level of confidence of 95\% ($\alpha = 0.05$) \cite{Demvsar-2006}. The results of the Nemenyi test are shown in Figure \ref{fig:nemenyiacc}.

As in the clustering experiment, the figure shows that using a parameter $r=1.5$ for the Tsallis regularization tends to yield the best results out of the three values of $r$ that were investigated.
Furthermore, the Wilcoxon signed-rank tests show that the FETsallis and the RSPTsallis perform significantly better than the SP, the Katz and the Modularity (except for the FETsallis with a parameter $r = 3$). The tests also show that the RSPTsallis with a parameter $r = 1.5$ obtains significantly better results than the LF. Nevertheless, except for the FETsallis with a parameter $r = 3$, no significant difference can be observed, in terms of ranks, between the KL and the Tsallis divergence regularization for both the FE and the RSP.

This confirms that the Tsallis RSP and FE dissimilarity measures, and especially the RSP, are able to capture the community structure of the graph in an accurate way, at least on the investigated datasets.

\begin{table}[t!]
\footnotesize
\begin{center}
\scalebox{0.75}{
\begin{tabular}{lccccccc}
\hline
\textbf{Classif. method $\rightarrow$}  & FE             & FETsallis(1.5) & FETsallis(2)   & FETsallis(3)   & RSP            & RSPTsallis(1.5) & RSPTsallis(2) \\
\textbf{Dataset} $\downarrow$             &                &                &                &                &           &&\\ \hline
IMDB                     & 75.31	&75.57	&77.18	&78.09	&76.07	&76.47	&76.55	\\
Newsgroup\_2\_1          & 96.79	&96.04	&95.44	&95.06	&95.96	&95.75	&95.56	 \\
Newsgroup\_2\_2          &91.21	&92.60	&93.10	&92.55	&91.22	&92.81	&\textbf{93.13}	 \\
Newsgroup\_2\_3          &95.98	&96.07	&96.30	&96.15	&95.39	&96.18	&96.35	 \\
Newsgroup\_3\_1          &92.53	&92.92	&92.86	&93.18	&92.88	&92.94	&92.83	 \\
Newsgroup\_3\_2          & 93.45	&93.19	&92.83	&92.65	&93.34	&\textbf{93.57}	&93.07 \\
Newsgroup\_3\_3          & \textbf{93.66}	&93.35	&92.29	&91.83	&92.91	&93.53	&93.03	 \\
Newsgroup\_5\_1          &\textbf{88.70	}&88.54	&88.24	&87.62	&88.09	&88.22	&88.03 \\
Newsgroup\_5\_2          & 82.32	&\textbf{82.55}	&82.38	&80.83	&81.49	&82.24	&81.23\\
Newsgroup\_5\_3          &80.27	&82.69	&82.78	&77.53	&80.38	&82.65	&\textbf{83.42}	 \\
WebKB-Cornell            & 57.20	&57.37	&56.52	&54.44	&57.53	&58.80	&58.31	 \\
WebKB-Texas              &73.35	&74.34	&72.32	&69.79	&75.24	&74.96	&72.82	\\
WebKB-Washington         & 68.15	&66.53	&66.26	&63.85	&68.88	&68.40	&66.15\\
WebKB-Wisconsin          &74.33	&70.69	&68.58	&68.66	&73.91	&71.67	&69.04 \\ \hline
\hline
\textbf{Classif. method $\rightarrow$}  & RSPTsallis(3)  & SP    & Modularity & Katz & LogCom & LF & SCT\\
\textbf{Dataset} $\downarrow$             &                &                &                &                &                &                 &             \\ \hline
IMDB                     	&\textbf{78.53}	&74.68	&74.37	&68.93	&76.28	&73.87	&78.36 \\
Newsgroup\_2\_1         	&94.51	&93.58	&95.85	&95.15	&96.18	&96.65	&\textbf{97.14} \\
Newsgroup\_2\_2         	&92.79	&90.36	&91.22	&90.38	&91.18	&90.36	&91.14 \\
Newsgroup\_2\_3         	&96.32	&\textbf{96.78}	&95.78	&93.21	&95.43	&95.54	&95.79 \\
Newsgroup\_3\_1         	&92.54	&93.01	&93.02	&91.00	&\textbf{94.00}	&92.98	&93.99 \\
Newsgroup\_3\_2        	&92.85	&89.32	&92.63	&91.25	&92.17	&92.15	&92.35 \\
Newsgroup\_3\_3         	&92.52	&91.14	&91.20	&88.73	&91.49	&90.93	&93.35 \\
Newsgroup\_5\_1         	&87.59	&86.60	&77.04	&79.74	&86.30	&86.22	&87.29 \\
Newsgroup\_5\_2         	&80.27	&78.36	&75.97	&64.47	&79.04	&80.10	&80.45 \\
Newsgroup\_5\_3          	&82.31	&72.73	&76.51	&66.33	&78.65	&79.94	&80.38 \\
WebKB-Cornell           	&55.62	&47.36	&50.71	&52.33	&\textbf{59.16}	&58.35	&57.30 \\
WebKB-Texas             	&72.93	&61.20	&73.01	&67.89	&\textbf{75.45}	&74.18	&74.27 \\
WebKB-Washington        	&64.69	&52.35	&62.52	&64.56	&\textbf{69.92}	&67.47	&67.66 \\
WebKB-Wisconsin         	&69.64	&62.46	&73.42	&73.61	&\textbf{74.99}	&74.48	&72.77  \\ \hline
\end{tabular}}
\end{center}
\caption{\footnotesize{Classification accuracies in percent for the various classification methods obtained on the different datasets. For each dataset and method, the final accuracy is obtained by averaging over 5 repetitions of a standard cross-validation procedure.  Each repetition consists of a nested cross-validation with 5 external folds (test sets, for validation) on which the accuracy of the classifier is averaged, and 5 internal folds (for parameter tuning). The best performing method is highlighted in boldface for each dataset.}}
\label{Tab:Acc}
\end{table}

\section{Conclusion and future work}
\label{Sec_conclusion01}

This paper showed that sparse randomized routing policies in a network can be obtained when regularizing the least-cost routing by the Tsallis divergence instead of the Kullback-Leibler divergence. Two different algorithms are detailed, a simple and faster procedure for the case $r = 2$ based on a linear search and a slower one for the more general case $r >1$ based on a bisection search technique. Indeed, in practice, we observed that the bisection method is significantly slower than the linear search method, at least on the investigated datasets.

In that context, various interesting quantities can be derived from the routing policy, especially the expected cost and the minimized free energy between the source and the target nodes. These quantities can be used as dissimilarity measures between the nodes of the network for tackling pattern recognition and machine learning tasks. A nice property is that they interpolate between the shortest path distance (when $\theta \rightarrow \infty$) and the commute-cost distance (proportional to the resistance distance, $\theta \rightarrow 0^{+}$). Another interesting property is the fact that, as the standard randomized shortest paths, they are taking the degree of inter-connectivity (direct and indirect), in addition to proximity, into account in the computation of the dissimilarity.

Indeed, it is well-known that the standard shortest path distance and the resistance distance \cite{Klein-1993}, while very useful in many contexts, show important drawbacks in some situations which hinders their use as distance measures between nodes in some applications. More precisely, the shortest path distance does not integrate the concept of high connectivity between the two nodes (it only considers the shortest paths, see, e.g., \cite{Fouss-2016}), while the resistance distance provides useless results when dealing with large graphs (the so-called ``lost-in-space effect" \cite{Luxburg-2010,Luxburg-2014}).
Another drawback of the shortest path distance is that it usually provides a large amount of ties when comparing distances, especially on unweighted and undirected graphs.
Moreover, it has been shown recently \cite{Hashimoto-2015} that the FE distance based on Kullback-Leibler regularization (called the logarithmic Laplace transformed hitting time in their paper) avoids to a certain extend the ``lost-in-space effect". Therefore, we conjecture that the FE dissimilarity based on the Tsallis divergence introduced in this paper benefits from the same property.

Experimental comparisons based on two pattern recognition tasks show that the proposed distances are competitive with other state-of-the-art techniques. The main drawback, however, is the fact that the computation of the dissimilarities in the general $r \ne 2$ case is time-consuming, preventing its application on large graphs.

Further work will be devoted to the improvement of the algorithm computing the FE distance based on the Tsallis divergence in the $r \ne 2$ case. For instance, we could adopt a mixed strategy by first applying the line search in order to identify the one-unit integer interval in which the optimal value lies, and then running a bisection search within this interval. Another idea would be to use the algorithm proposed in \cite{Duchi-2008} for computing the orthogonal projection on the unit simplex, based on a modification of the randomized median finding procedure \cite{Blum-1973,Cormen-2009}. The link between the orthogonal projection on the unit simplex and our formulation (\ref{Eq_sparse_optimization_problem_reference01}) should also be studied.

Still another interesting contribution would be to apply the Tsallis divergence regularization for solving Markov decision problems, inducing sparse policies; therefore extending the work of \cite{Lee-2018}.
We also plan to use the Tsallis divergence for the design of algorithms solving the sparse randomized optimal transport on a graph problem, extending previous work \cite{Guex-2019}. Indeed, in the same way as in this paper, the Kullback-Leibler divergence can be replaced by the Tsallis divergence as regularization term in the model, therefore providing sparse routing policies.

\section*{Acknowledgements}

\noindent This work was partially supported by the Immediate and the Brufence projects funded by InnovIris (Brussels Region), as well as former projects funded by the Walloon region, Belgium. We thank these institutions for giving us the opportunity to conduct both fundamental and applied research. We also thank Professor Masashi Shimbo and Dr Amin Mantrach for the helpful references and discussions.

\begin{center}
\rule{2.5in}{0.01in}
\end{center}


\appendix
\section*{Appendices}
\label{Appendix}

These appendices discuss the alternative form of the randomized shortest paths, the convexity of the objective function with Tsallis divergence regularization, as well as the algorithms for computing the $\mathrm{spmin}$ function appearing in Equation (\ref{Eq_sparse_min_reference01}), returning transition probabilities $\mathbf{p}_{i}$ associated to a node $i$.
To the best of our knowledge, these algorithms were first studied in \cite{Kanzawa-2013,Miyamoto-1998} (although we suspect that they have probably been investigated before). This appendix provides a reformulation of the relevant material contained in these papers (\ref{Appendix_B1} -- \ref{Appendix_B2}), as well as an extension of their algorithms for dealing with an arbitrary reference distribution (and thus regularizing with Tsallis divergence instead of Thsallis entropy, \ref{Appendix_B3}).

\section{An alternative view of the standard randomized shortest path framework}
\label{Ap_alternative_randomized_shortest_paths01}

The path-based formalism (\ref{Eq_optimization_problem_BoP01}) can be transformed into a ``local" form (see \cite{Akamatsu-1997,Garcia-Diez-2011,Saerens-2008}) which will be used for deriving the sparse RSP. In this new form, the policy is computed in an iterative way by exploiting Lagrange duality.

\subsection{Alternative form of the objective function}
\label{Subsec_alternative_form_RSP01}

To this end, let us introduce $\eta\big( (i,j) \in \wp \big)$ defined as the number of times edge $(i,j)$ is visited along path $\wp$. Because the probability of a path can be expressed as a product of transition probabilities (see the discussion after Equation (\ref{Eq_path_transition_probabilities01})), the path-based quantities $\log (\mathrm{P}(\wp)/\tilde{\pi}(\wp))$ and $\tilde{c}(\wp)$ can be expressed as
\begin{equation}
\begin{cases}
\log \dfrac{ \mathrm{P}(\wp) } { \tilde{\pi}(\wp) } = \dsum_{(i,j) \in \mathcal{E}} \eta\big( (i,j) \in \wp \big) \log \dfrac{ p_{ij} } { p^{\mathrm{ref}}_{ij} } \\
\tilde{c}(\wp) = \dsum_{(i,j) \in \mathcal{E}} \eta\big( (i,j) \in \wp \big) \, c_{ij}
\end{cases}
\label{Eq_total_cost_from_eta01}
\end{equation}
Injecting these relations in the objective function appearing in Equation (\ref{Eq_optimization_problem_BoP01}) yields
\begin{align}
\phi^{\mathrm{\textsc{kl}}}_{st}(\mathrm{P}) &= \dsum_{\wp \in \mathcal{P}_{st}} \mathrm{P}(\wp) \tilde{c}(\wp) + T \dsum_{\wp \in \mathcal{P}_{st}} \mathrm{P}(\wp) \log \left( \frac{\mathrm{P}(\wp)}{\tilde{\pi}(\wp)} \right) \nonumber \\
&= \dsum_{\wp \in \mathcal{P}_{st}} \mathrm{P}(\wp) \dsum_{(i,j) \in \mathcal{E}} \eta\big( (i,j) \in \wp \big) \, c_{ij} + T \dsum_{\wp \in \mathcal{P}_{st}} \mathrm{P}(\wp) \dsum_{(i,j) \in \mathcal{E}} \eta\big( (i,j) \in \wp \big) \log \dfrac{ p_{ij} } { p^{\mathrm{ref}}_{ij} } \nonumber \\
&= \dsum_{(i,j) \in \mathcal{E}} \bar{n}_{ij} \, c_{ij} + T \dsum_{(i,j) \in \mathcal{E}} \bar{n}_{ij} \log \dfrac{ p_{ij} } { p^{\mathrm{ref}}_{ij} } \nonumber \\
&= \dsum_{(i,j) \in \mathcal{E}} \bar{n}_{i} p_{ij} \bigg( c_{ij} + T \log \dfrac{ p_{ij} } { p^{\mathrm{ref}}_{ij} } \bigg)
\label{Eq_local_formulation_objective_function01}
\end{align}
 where $\bar{n}_{ij} \triangleq \sum_{\wp \in \mathcal{P}_{st}} \mathrm{P}(\wp) \, \eta\big( (i,j) \in \wp \big)$ is the expected number of passages (the flow) through edge $(i,j)$ and we used $\bar{n}_{ij} = \bar{n}_{i} p_{ij}$ with $\bar{n}_{i} = \bar{n}_{i \bullet} = \sum_{j \in \mathcal{S}ucc(i)} \bar{n}_{ij}$ denoting the expected number of visits to node $i$. This comes from the fact that the expected  number of passages through edge $(i,j)$ is equal to the number of visits to node $i$ times the probability of following the link $(i,j)$ from node $i$. This formulation of the RSP problem is also closely related to the framework of \cite{Bavaud-2012,Guex-2015} based on edge flows and flow conservation.
 
The objective function (\ref{Eq_local_formulation_objective_function01}) should be minimized with respect to the (local) policy, that is, the set of transition probabilities $p_{ij}$ associated to edges. It is shown in \cite{Akamatsu-1997} that this objective function is strictly convex with respect to edge flows, $\bar{n}_{ij} = \bar{n}_{i} p_{ij}$. Indeed, the objective function, which is expressed in (\ref{Eq_local_formulation_objective_function01}) in terms of transition probabilities and number of visits to nodes, can also be expressed in function of edge flows only, or in function of transition probabilities only. Therefore, because the correspondence between edge flows and transition probabilities is differentiable and one-to-one for a unit input flow (which is indeed the case in the RSP model), any stationary point of (\ref{Eq_local_formulation_objective_function01}) with respect to the transition probabilities $p_{ij}$ is also a stationary point with respect to the corresponding edge flows .
Thus, because the objective function is convex with respect to the $\bar{n}_{ij}$ and the domain is convex, it must be a global minimum.

Interestingly, this also shows that the path-based formalism of Equation (\ref{Eq_optimization_problem_BoP01}) is equivalent to minimizing the local cost plus KL divergence, $c_{ij} + T \log ( p_{ij} / p^{\mathrm{ref}}_{ij} )$, which is also the purpose of Kullback-Leibler, or path integral, control developed in the field of reinforcement learning and control theory. Therefore, as already mentioned in the related work (Subsection \ref{Subsec_related_work01}), the randomized shortest paths framework is equivalent to some of these models developed in reinforcement learning  (see, e.g., \cite{Busic-2018,Fox-2016,Kappen-2012,Rubin-2012,Theodorou-2013,Theodorou-2012}), as initiated by \cite{Todorov-2007,Todorov-2008}.

\subsection{Computing the optimal policy}
\label{Ap_optimal_policy_computation01}

In this subsection, an algorithm for computing the optimal transition probabilities (the policy) is developed. It aims at minimizing the objective function (\ref{Eq_local_formulation_objective_function01}) by considering the transition probabilities and the expected number of visits as independent. The dependency between the two quantities is introduced as a constraint in the formulation, as commonly done in discrete-state discrete-time optimal control (see, e.g., \cite{Luenberger-1979}).
After \emph{renumbering the nodes} in such a way that node $1$ is the source node and node $n$ the target node\footnote{It is assumed that the source node is different from the target node.} for convenience, this leads to the following Lagrange function only including the equality constraints
\begin{align}
\mathscr{L}(\mathbf{P},\bar{\mathbf{n}};\boldsymbol{\mu},\boldsymbol{\lambda}_{\mathrm{\textsc{kl}}})
&= \dsum_{i \in \mathcal{V} \setminus n} \bar{n}_{i} \dsum_{j \in \mathcal{S}ucc(i)} p_{ij}  \bigg( c_{ij} + T \log \dfrac{ p_{ij} } { p^{\mathrm{ref}}_{ij} } \bigg)  \nonumber \\
&\quad + \dsum_{i \in \mathcal{V} \setminus n} \mu_{i} \bigg( 1 - \dsum_{j \in \mathcal{S}ucc(i)} p_{ij} \bigg) \nonumber \\
&\quad + \dsum_{j \in \mathcal{V}} \lambda^{\mathrm{\textsc{kl}}}_{j} \bigg( \dsum_{i \in \mathcal{P}red(j)} \bar{n}_{i} p_{ij} + \delta_{1j} - \bar{n}_{j} \bigg)
\label{Eq_local_formulation_lagrange_function01}
\end{align}
where $\mathcal{S}ucc(i)$ is the set of successor nodes\footnote{Recall that target node $n$ is killing and absorbing, and thus has no successor. It is therefore not the predecessor of any node.} of node $i$ and $\mathcal{P}red(j)$ is the set of predecessor nodes of node $j$. The quantities $\mu_{i}$, $\lambda^{\mathrm{\textsc{kl}}}_{i}$ are standard Lagrange parameters dealing with equality constraints. The transition probabilities and the expected number of visits are therefore considered as independent in the optimization of the Lagrange function and must be non-negative. The relation between $\bar n_{i}$ and $p_{ij}$ is given by the system of linear equations computing the expected number of visits to nodes in an absorbing Markov chain, $\bar{n}_{j} = \sum_{i \in \mathcal{P}red(j)} \bar{n}_{i} p_{ij} + \delta_{1j}$ for each $j \in \mathcal{V}$, when a unit flow is injected in node 1 (see, e.g., \cite{Norris-1997,Taylor-1998}). By the property of flow conservation in a Markov chain, it is clear that $\bar{n}_{n} = 1$ for the target node.


Our procedure optimizes sequentially the objective function by Lagrange duality as follows \cite{Beck-2014,Culioli-2012,Griva-2008,Minoux-1986}. We first minimize the Lagrange function with respect to the transition probabilities $p_{ij}$ subject to sum-to-one constraints, while fixing the Lagrange parameters $\lambda^{\mathrm{\textsc{kl}}}_{i}$. We will observe that they only depend on the Lagrange parameters. Lagrange parameters are then computed by maximizing the dual, a common optimization procedure called the Arrow-Hurwicz-Uzawa algorithm \cite{Arrow-1958}. The two steps are iterated until convergence, which is guaranteed because each sub-problem reaches its optimum uniquely \cite{Bertsekas-1999}. In practice, we observed that the duality gap is always zero, showing that a global minimum is reached.

\subsubsection{Computation of the transition probabilities}
\label{Ap_subsubsec_Shannon-transition01}

For the estimation of the transition probabilities step, there is no need to introduce non-negativity constraints because KL divergence regularization ensures that the estimates satisfy the constraint \cite{Kapur-1989}.
Taking the partial derivative of the Lagrange function (\ref{Eq_local_formulation_lagrange_function01}) with respect to the transition probabilities associated to edges and setting the result to zero provides
\begin{equation}
T \log \frac{ p_{ij} } { p_{ij}^{\mathrm{ref}} } = \frac{ \mu_{i} } { \bar{n}_{i} } - T - (c_{ij} + \lambda^{\mathrm{\textsc{kl}}}_{j}) \nonumber
\end{equation}
Then, using $\theta = 1/T$, isolating the transition probabilities as well as imposing the sum-to-one constraint yields
\begin{equation}
 p_{ij} = \frac{ p_{ij}^{\mathrm{ref}} \exp[-\theta (c_{ij} + \lambda^{\mathrm{\textsc{kl}}}_{j})] } { \dsum_{k \in \mathcal{S}ucc(i)} p_{ik}^{\mathrm{ref}} \exp[-\theta (c_{ik} + \lambda^{\mathrm{\textsc{kl}}}_{k})] } \quad \text{for all } (i,j) \in \mathcal{E}
 \label{Eq_transition_probabilities_computation01}
\end{equation}
which only depends on the Lagrange parameters $\lambda^{\mathrm{\textsc{kl}}}_{i}$.
This corresponds to the ``local" optimal randomized policy for going from $1$ to $n$, according to KL divergence regularization.
%
Let us now compute these Lagrange parameters.

\subsubsection{Computation of the Lagrange parameters}
\label{Ap_subsubsec_Shannon-Lagrange01}

%
Computing the $\lambda^{\mathrm{\textsc{kl}}}_{i}$ aims at solving the dual problem.
Indeed, by defining respectively the expected cost and the KL divergence per node, $\tilde{c}_{i} = \sum_{j \in \mathcal{S}ucc(i)} p_{ij} c_{ij}$ and $\bar{h}_{i}^{\mathrm{\textsc{kl}}} = \sum_{j \in \mathcal{S}ucc(i)} p_{ij} \log ( p_{ij} / p^{\mathrm{ref}}_{ij})$ for $i \ne n$, together with $\bar{c}_{n} = 0$ and $\bar{h}_{n}^{\mathrm{\textsc{kl}}} = 0$ for target node, the problem of computing the expected number of visits $\bar{n}_{i}$ when transition probabilities are fixed can be reformulated from (\ref{Eq_local_formulation_lagrange_function01}) as a linear programming problem: minimize $(\bar{\mathbf{c}} + T \bar{\mathbf{h}}_{\mathrm{\textsc{kl}}})^{\mathrm{T}} \bar{\mathbf{n}}$ with respect to $\bar{\mathbf{n}}$ subject to the constraints $(\mathbf{I} - \mathbf{P})^{\mathrm{T}} \bar{\mathbf{n}} = \mathbf{e}_{1}$ and $\bar{\mathbf{n}} \ge \mathbf{0}$.
%

However, because we instead need the vector of Lagrange parameters $\boldsymbol{\lambda}_{\mathrm{\textsc{kl}}}$ in order to compute the transition probabilities (see (\ref{Eq_transition_probabilities_computation01})), we are more interested in the dual problem (see, e.g., \cite{Griva-2008}, page 184\footnote{Note that the equation in \cite{Griva-2008} also holds for equality constraints, providing $(\mathbf{I} - \mathbf{P})^{\mathrm{T}} \bar{\mathbf{n}} = \mathbf{e}_{1}$, which is the case here.}),
\begin{equation}
\vline\,\begin{array}{llll}
\maximize\limits_{\boldsymbol{\lambda}_{\mathrm{\textsc{kl}}}} & \mathbf{e}_{1}^{\mathrm{T}} \boldsymbol{\lambda}_{\mathrm{\textsc{kl}}} \\[0.3cm]
\subjectto & (\mathbf{I} - \mathbf{P}) \boldsymbol{\lambda}_{\mathrm{\textsc{kl}}} = \bar{\mathbf{c}} + T \bar{\mathbf{h}}_{\mathrm{\textsc{kl}}} \\
& \boldsymbol{\lambda}_{\mathrm{\textsc{kl}}} \ge \mathbf{0}
\end{array}
\label{Eq_dual_problem01}
\end{equation}
where $\mathbf{e}_{1}$ is a column vector full of $0$'s, except in position $1$ containing a $1$.
Because the elements of $(\mathbf{I} - \mathbf{P})^{-1} = \mathbf{I} + \mathbf{P} + \mathbf{P}^{2} + \cdots$ are all non-negative, the non-negativity constraint on the Lagrange parameters is automatically satisfied. These Lagrange parameters are thus obtained by solving the system of linear equations
\begin{equation}
(\mathbf{I} - \mathbf{P}) \boldsymbol{\lambda}_{\mathrm{\textsc{kl}}} = \bar{\mathbf{c}} + T \bar{\mathbf{h}}_{\mathrm{\textsc{kl}}}
\label{Eq_free_energy_computation01}
\end{equation}
where, in matrix form, $\bar{\mathbf{c}} = (\mathbf{P} \circ \mathbf{C}) \mathbf{e}$ and $\bar{\mathbf{h}}_{\mathrm{\textsc{kl}}} = (\mathbf{P} \circ (\log \mathbf{P} - \log \mathbf{P}_{\mathrm{ref}})) \mathbf{e}$, with $\mathbf{e}$ being a column vector of $1$'s and $\circ$ the elementwise matrix product. Elementwise, we have
\begin{equation}
\lambda^{\mathrm{\textsc{kl}}}_{i} = \sum_{j \in \mathcal{S}ucc(i)} p_{ij} \bigg( c_{ij} + \lambda^{\mathrm{\textsc{kl}}}_{j} + T \log \frac{ p_{ij} } { p_{ij}^{\mathrm{ref}} } \bigg)
\label{Eq_free_energy_elementwise_computation01}
\end{equation}
and notice that this implies $\lambda^{\mathrm{\textsc{kl}}}_{n} = 0$ for the target node $n$.
The procedure aims at iterating Equations (\ref{Eq_transition_probabilities_computation01}) and (\ref{Eq_free_energy_computation01}) until convergence. Interestingly, the Lagrange parameters have a nice interpretation, as explained in the next subsection.

\subsubsection{Interpretation of the Lagrange parameters}

Let us now give an interpretation to the Lagrange parameters $\boldsymbol{\lambda}_{\mathrm{\textsc{kl}}}$. From Equation (\ref{Eq_transition_probabilities_computation01}), we directly obtain
\begin{equation}
T \log \frac{ p_{ij} } { p_{ij}^{\mathrm{ref}} } = - (c_{ij} + \lambda^{\mathrm{\textsc{kl}}}_{j})
- T \log \dsum_{k \in \mathcal{S}ucc(i)} p_{ik}^{\mathrm{ref}} \exp[-\theta (c_{ik} + \lambda^{\mathrm{\textsc{kl}}}_{k})] \nonumber
\end{equation}
By injecting this expression in Equation (\ref{Eq_free_energy_elementwise_computation01}), we obtain
\begin{equation}
\lambda^{\mathrm{\textsc{kl}}}_{i} = - \tfrac{1}{\theta} \log \dsum_{j \in \mathcal{S}ucc(i)} p_{ij}^{\mathrm{ref}} \exp[-\theta (c_{ij} + \lambda^{\mathrm{\textsc{kl}}}_{j})]
\label{Eq_BellmanFord_Lagrange01}
\end{equation}
which is nothing else than the Bellman-Ford-like recurrence formula\footnote{The min operator in the Bellman-Ford expression is replaced by a softmin operator, see \cite{Francoisse-2017}.} for computing the directed \emph{free energy distance} $\boldsymbol{\phi}^{\mathrm{\textsc{kl}}}$ (see\footnote{As it plays the role of a potential, the FE distance was called the potential distance in \cite{Francoisse-2017}.} \cite{Francoisse-2017}, Eq. (34)); therefore $\boldsymbol{\lambda}^{\mathrm{\textsc{kl}}} = \boldsymbol{\phi}^{\mathrm{\textsc{kl}}}$ and the Lagrange parameters are equal to the directed FE distances at the optimum.

This directed FE distance is the minimum free energy obtained by replacing path probabilities by the optimal ones (see Equation (\ref{Eq_Boltzmann_probability_distribution01})) in the free energy objective function provided by Equation (\ref{Eq_optimization_problem_BoP01}) (see \cite{Kivimaki-2012} for details). This quantity has many interesting properties: (1) it plays the role of a potential at the continuous space-time limit \cite{Garcia-Diez-2011b}, (2) it defines a distance measure between nodes when symmetrized \cite{Francoisse-2017,Kivimaki-2012}, (3) it interpolates between the least-cost and the commute-cost distance, (4) it can be interpreted as minus $T$ times the log-likelihood of surviving during a particular killed random walk \cite{Francoisse-2017}, and (5) it performed consistently well in a number of pattern recognition tasks \cite{Francoisse-2017,Sommer-2016,Sommer-2017}, among others.

\section{About the convexity of the Tsallis-regularized objective function}
\label{Ap_convexity_objective_function01}

This section discusses the convexity of the Tsallis-regularized free energy objective function appearing in Equation (\ref{Eq_primal_problem_Tsallis01}). To this end, we will compute the Hessian matrix and verify if the corresponding quadratic form is positive semi-definite.

Using $\bar{n}_{ij} = \bar{n}_{i} p_{ij}$ with $\bar{n}_{i} = \bar{n}_{i \bullet} = \sum_{j \in \mathcal{S}ucc(i)} \bar{n}_{ij}$, we first reformulate this objective function in terms of edge flows $\bar{n}_{ij}$,
\begin{align}
\phi^{\mathrm{\textsc{t}s}}_{st} 
&= \dsum_{i \in \mathcal{V}} \bar{n}_{i \bullet} \dsum_{j \in \mathcal{S}ucc(i)} p_{ij} \bigg( c_{ij} + \tfrac{T}{r-1} \bigg( \bigg( \dfrac{ p_{ij} } { p^{\mathrm{ref}}_{ij} } \bigg)^{\hspace{-3pt} r-1} - 1 \bigg) \bigg) \nonumber \\
&= \dsum_{i \in \mathcal{V}} \dsum_{j \in \mathcal{S}ucc(i)} \bigg( c_{ij} \bar{n}_{ij} + \tfrac{T}{r-1}  \dfrac{ (\bar{n}_{ij})^{r} } { (p^{\mathrm{ref}}_{ij} \bar{n}_{i \bullet})^{r-1} } - \tfrac{T}{r-1} \bar{n}_{i j} \bigg)
\end{align}

For computing the Hessian, only the central term is meaningful. We will therefore study the function (at first, we do not consider the reference probability $p^{\mathrm{ref}}_{ij}$ for simplicity),
\begin{equation}
   f(\mathbf{N}) = \dsum_{k \in \mathcal{V}} \dsum_{l \in \mathcal{S}ucc(k)} \dfrac{ (\bar{n}_{kl})^{r} } { \hspace{+5pt}( \bar{n}_{k \bullet})^{r-1} }
\end{equation}

The first-order partial derivatives are
\begin{equation}
    \dfrac{\partial f}{\partial \bar{n}_{i j}} = r \, p_{ij}^{r-1} - (r-1) {\textstyle \sum_{j' \in \mathcal{S}ucc(i)} } p_{ij'}^{r}
\end{equation}
where $p_{ij}$ denotes $\bar{n}_{i j} / \bar{n}_{i \bullet}$ for simplicity.
Then the Hessian is
\begin{equation}
    h_{(i,j)(k,l)} = \dfrac{\partial^{2} f}{\partial \bar{n}_{k l} \partial \bar{n}_{i j}}
    = r (r-1) \frac{\delta_{ik}}{\bar{n}_{i \bullet}} \Big( \delta_{jl} p_{ij}^{r-2} - (p_{ij}^{r-1} + p_{il}^{r-1}) + {\textstyle \sum_{j' \in \mathcal{S}ucc(i)} } p_{ij'}^{r} \Big)
\end{equation}

Thus, the following quadratic form should be non-negative for any value of $x_{(i,j)}$,
\begin{align}
    Q &= \sum_{(i,j) \in \mathcal{E}} \sum_{(k,l) \in \mathcal{E}} x_{(i,j)} h_{(i,j)(k,l)} x_{(k,l)} \nonumber \\
    &= r (r-1) \sum_{i \in \mathcal{V}} \frac{1}{\bar{n}_{i \bullet}} \dsum_{j,l \in \mathcal{S}ucc(i)} x_{(i,j)} \underbracket[0.5pt][3pt]{ \Big( \delta_{jl} p_{ij}^{r-2} - (p_{ij}^{r-1} + p_{il}^{r-1}) + {\textstyle \sum_{j' \in \mathcal{S}ucc(i)} } p_{ij'}^{r} \Big) }_{\text{matrix } q_{jl}(i)} x_{(i,l)}
\end{align}
and therefore it suffices to show that the matrices $\mathbf{Q}(i) = (q_{jl}(i))$ are positive semi-definite in order to prove convexity ($\bar{n}_{i \bullet}$ is always positive).
When introducing the reference probabilities, the objective function becomes
\begin{equation}
   f(\mathbf{N}) = \dsum_{k \in \mathcal{V}} \dsum_{l \in \mathcal{S}ucc(k)} (p^{\mathrm{ref}}_{kl})^{1-r} \dfrac{ (\bar{n}_{kl})^{r} } { \hspace{+5pt}( \bar{n}_{k \bullet})^{r-1} }
\end{equation}
%
and the Hessian is
\begin{align}
    h_{(i,j)(k,l)}
    &= r (r-1) \frac{\delta_{ik}}{\bar{n}_{i \bullet}} \Big( \delta_{jl} (p^{\mathrm{ref}}_{ij})^{1-r} \, p_{ij}^{r-2} - \big( (p^{\mathrm{ref}}_{ij})^{1-r} \, p_{ij}^{r-1} + (p^{\mathrm{ref}}_{il})^{1-r} \, p_{il}^{r-1} \big) \nonumber \\
    &\quad + {\textstyle \sum_{j' \in \mathcal{S}ucc(i)} } (p^{\mathrm{ref}}_{ij'})^{1-r} \, p_{ij'}^{r} \Big)
\end{align}

Then, the quadratic form can be readily deduced
\begin{align}
    Q
    &= r (r-1) \sum_{i \in \mathcal{V}} \frac{1}{\bar{n}_{i \bullet}} \dsum_{j,l \in \mathcal{S}ucc(i)} x_{(i,j)} \Big( \delta_{jl} (p^{\mathrm{ref}}_{ij})^{1-r} \, p_{ij}^{r-2} - \big( (p^{\mathrm{ref}}_{ij})^{1-r} \, p_{ij}^{r-1} + (p^{\mathrm{ref}}_{il})^{1-r} \, p_{il}^{r-1} \big) \nonumber \\
    &\quad + {\textstyle \sum_{j' \in \mathcal{S}ucc(i)} } (p^{\mathrm{ref}}_{ij'})^{1-r} \, p_{ij'}^{r} \Big) x_{(i,l)}
\end{align}
Because we did not find an easy way to prove formally the positive semi-definiteness of the matrices $\mathbf{Q}(i)$ with elements $q_{jl}(i) = \delta_{jl} (p^{\mathrm{ref}}_{ij})^{1-r} \, p_{ij}^{r-2} - \big( (p^{\mathrm{ref}}_{ij})^{1-r} \, p_{ij}^{r-1} + (p^{\mathrm{ref}}_{il})^{1-r} \, p_{il}^{r-1} \big) + {\textstyle \sum_{j' \in \mathcal{S}ucc(i)} } (p^{\mathrm{ref}}_{ij'})^{1-r} \, p_{ij'}^{r} $, we decided to test it numerically. More precisely, we generated $10^{6}$ random instances of 20-dimensional probability vectors $\mathbf{p}_{i}$ and $\mathbf{p}^{\mathrm{ref}}_{i}$, as well as values of the $r$ parameter in $[1.1, 4.1]$. The smallest eigenvalue of the corresponding $\mathbf{Q}(i)$ matrix is then extracted. In all cases, the smallest eigenvalue was equal to $\lambda_{\mathrm{min}} = 0$ (no negative eigenvalue), up to small errors of the order $| \lambda_{\mathrm{min}} / \lambda_{\mathrm{max}} | < 10^{-14}$. This provides evidence that the objective function is convex, although it remains a conjecture at this point.

\section{Minimizing expected cost plus Tsallis free energy}
\label{Appendix_B}

In this section, we solve the problem stated in Equation (\ref{Eq_sparse_optimization_problem_reference01}), that is, the minimization of an expected cost with Tsallis divergence regularization. We proceed gradually in three steps. First, the $r=2$ case and a uniform reference probability distribution is considered (\ref{Appendix_B1}). Then, we extend the results to the more general $r > 1$ case (\ref{Appendix_B2}). Finally, the most general case of $r > 1$ and a non-uniform reference probability distribution (and thus Tsallis divergence regularization) of Equation (\ref{Eq_sparse_optimization_problem_reference01}) is considered  (\ref{Appendix_B3}). Note that \ref{Appendix_B1}-\ref{Appendix_B2} are based on \cite{Kanzawa-2013,Miyamoto-1998}.

\subsection{Results for the $r=2$ case and a uniform reference probabilities}
\label{Appendix_B1}

This subsection derives the algorithm for finding a sparse solution to the problem of minimizing an expected cost under quadratic constraints, thus in the case where $r=2$ and a uniform reference distribution (adapted from \cite{Kanzawa-2013,Miyamoto-1998}). The problem stated in Equation (\ref{Eq_sparse_optimization_problem_reference01}) then reduces to
\begin{equation}
\vline\,\begin{array}{llll}
\minimize\limits_{\mathbf{p}} &  \mathbf{c}^{\mathrm{T}} \mathbf{p}  +  T \, \| \mathbf{p}  \|^{2}_{2}   \\[0.3cm]
\subjectto & \mathbf{e}^{\mathrm{T}} \mathbf{p} = 1 \\
           & \mathbf{p} \ge \mathbf{0}
\end{array}
\label{Eq_simple_r_2_case01}
\end{equation}
where $T$ is the temperature parameter, and it is assumed that vectors $\mathbf{p}$ and $\mathbf{c} \ge \mathbf{0}$ contain $m$ elements (corresponding to successor nodes of node $i$). We omit the row index $i$ as well as the quote for the augmented cost for the sake of simplicity. In our context, the vector $\mathbf{c}$ corresponds to the augmented costs, $\mathbf{c}'_{i}$, associated to the outgoing edges of a transient node $i$ and $\mathbf{p}$ corresponds to its transition probabilities $\mathbf{p}_{i}$ (see Equation (\ref{Eq_sparse_optimization_problem_reference01})).


\subsubsection{The Karush-Kuhn-Tucker conditions}

By using the necessary Karush-Kuhn-Tucker conditions \cite{Griva-2008,Luenberger-2010,Rardin-1998} and introducing the Lagrange parameters $\mu$ (equality constraint in (\ref{Eq_simple_r_2_case01})) and $\boldsymbol{\lambda}$ (inequality constraints in (\ref{Eq_simple_r_2_case01})), we obtain, in addition to the sum-to-one and non-negativity constraints on $\mathbf{p}$,
\begin{equation}
\begin{cases}
\mathbf{c} + 2 T \mathbf{p} - \mu \mathbf{e} - \boldsymbol{\lambda} = \mathbf{0} \\
\boldsymbol{\lambda}^{\mathrm{T}} \mathbf{p} = 0\\
\boldsymbol{\lambda} \ge \mathbf{0}
\end{cases}
\label{Eq_KKT01}
\end{equation}
from which we immediately deduce $\lambda_{i} p_{i} = 0$ for each $i$. In other words, $\lambda_{i} = 0$ or $p_{i} = 0$ because both $\boldsymbol{\lambda}$ and $\mathbf{p}$ are non-negative. We further denote by $\mathcal{Q}_{+}^{*}$ the set of \emph{strictly positive} $p_{i}$ (to be found) and $| \mathcal{Q}_{+}^{*} |$ the number of such elements.

Let us now consider the different cases.
First, if $p_{i} = 0$ for element $i$, Equation (\ref{Eq_KKT01}) tells us that $\lambda_{i} = c_{i} - \mu$. And because $\lambda_{i} \ge 0$, we must have $c_{i} \ge \mu$ when $p_{i} = 0$. By taking the contraposition of the previous implication and using the fact that $p_{i} \ge 0$, we obtain that if $c_{i} < \mu$ then $p_{i} > 0$ (and thus also $\lambda_{i} = 0$).

Next, we investigate the situation where $c_{i} \ge \mu$, equivalent to $\mu - c_{i} \le 0$. From (\ref{Eq_KKT01}), this implies $2 T p_{i} - \lambda_{i} \le 0$ and thus $p_{i} \le \lambda_{i} / 2 T$. Because the $p_{i}$ are non-negative and $\lambda_{i} = 0$ or $p_{i} = 0$, this implies that $p_{i} = 0$ when $c_{i} \ge \mu$.
The parameter $\mu$ is therefore a \emph{threshold} telling us when $p_{i}$ should be put to zero.

Finally if, for element $i$, $p_{i} > 0$ then $\lambda_{i} = 0$ and the non-negativity constraint on $p_{i}$ is non-active. Then, from (\ref{Eq_KKT01}), $c_{i} + 2  T p_{i} - \mu = 0$, which provides $p_{i} = \tfrac{1}{2 T} (\mu - c_{i})$.
Therefore, the $p_{i}$ are of the following form
\begin{equation}
p_{i} = \tfrac{1}{2 T} [\mu - c_{i}]_{+}
= \begin{cases}
\tfrac{1}{2 T} (\mu - c_{i}) & \text{\small{when }} \mu - c_{i} > 0 \\
0 & \text{\small{when }} \mu - c_{i} \le 0
\end{cases}
\label{Eq_KKT_result01}
\end{equation}
where $[x]_{+} = \max(x,0)$: if $x$ is negative, it is replaced by $0$. We observe that $p_{i}$ is put to zero when $(\mu - c_{i})$ becomes non-positive.
Therefore, without loss of generality, from now we assume a numbering such that the elements of the cost vector $\mathbf{c}$ are \emph{sorted and indexed by increasing value} of $c_{i}$, $c_{1} \le c_{2} \le \cdots \le c_{m}$.
In that case, from (\ref{Eq_KKT_result01}), the sequence of $( p_{i} )_{i=1}^{m}$ is monotonic non-increasing for a fixed $\mu$.

The result (\ref{Eq_KKT_result01}) also tells us that the optimal set of strictly positive $p_{i}$ is given by $\mathcal{Q}_{+}^{*} = \{ 1, 2, \dots, k^{*} \}$, for some index $k^{*} = | \mathcal{Q}_{+}^{*} |$ to be found. We now have to compute the Lagrange parameter $\mu$ as well as this threshold index $k^{*}$.

\subsubsection{Computing the Lagrange parameter $\mu$}

By expressing the sum-to-one constraint on $\mathbf{p}$ and assuming that the optimal number of strictly positive elements $| \mathcal{Q}_{+}^{*} | = k^{*}$ is known, the Lagrange parameter can easily be computed, $\mu = \tfrac{1}{| \mathcal{Q}_{+}^{*} |} ( \sum_{j \in \mathcal{Q}_{+}^{*}} c_{j} ) + \frac{2 T}{| \mathcal{Q}_{+}^{*} |}$.
By injecting this expression into Equation (\ref{Eq_KKT_result01}), we obtain for $i \in \mathcal{Q}_{+}^{*}$ and thus $i \le k^{*}$
\begin{equation}
p_{i} = \tfrac{1}{2 T} \bigg( \tfrac{1}{| \mathcal{Q}_{+}^{*} |} \sum_{j \in \mathcal{Q}_{+}^{*}} c_{j} - c_{i} \bigg) + \frac{1}{| \mathcal{Q}_{+}^{*} |}
= \tfrac{1}{2 T} \bigg( \tfrac{1}{k^{*}} \underbracket[0.5pt][3pt]{ \sum_{j=1}^{k^{*}} c_{j} }_{= s(k^{*})} - c_{i} \bigg) + \frac{1}{k^{*}}
\label{Eq_membership_function01}
\end{equation}
and this expression can be extended to the whole set of $p_{i}$, $i=1, \dots, m$, with
\begin{equation}
p_{i} = \bigg[ \tfrac{1}{2 T} \bigg( \tfrac{1}{k^{*}} \sum_{j=1}^{k^{*}} c_{j} - c_{i} \bigg) + \frac{1}{k^{*}} \bigg]_{+}
\label{Eq_membership_function_general01}
\end{equation}
where $[x]_{+} = \max(x,0)$.
When $T$ is large, we obtain a uniform distribution, $p_{i} = 1/k^{*}$ for $i = 1, \dots, k^{*}$, whereas when $T$ is close to zero, the probability mass is concentrated on the first element (the lowest cost), $p_{i} = \delta_{i1}$, a Kronecker delta.
Let us now compute the threshold index $k^{*}$.

\subsubsection{Computing the threshold index $k^{*}$}

The idea now for finding $k^{*}$ is to investigate sequentially a number of positive elements $k = 1,2, \dots$ for $\mathbf{p}$ and select the $k$ for which the L1 norm of vector $\mathbf{p}$ will be equal or greater than $1$.
More precisely, from Equation (\ref{Eq_KKT_result01}), assuming a number of positive elements $k \le k^{*}$ is selected, the sum of the corresponding entries of the vector $\mathbf{p}$ (the L1 norm) is equal to
\begin{equation}
\sum_{i = 1}^{k} p_{i} = \tfrac{1}{2 T} \sum_{i = 1}^{k} (\mu - c_{i}) = \tfrac{1}{2 T} \bigg( k \mu - \sum_{i = 1}^{k} c_{i} \bigg)
\label{Eq_L1_computation01}
\end{equation}
%
The optimal threshold $\mu$ is found when the L1 norm is \emph{exactly equal} to 1 (the only solution which is admissible). The idea is thus to increase $k$ until\footnote{Recall that the elements are sorted by increasing cost value.} the L1 norm is equal to or exceeds one, as proposed in \cite{Kanzawa-2013}.

To this end, let us introduce the following auxiliary function from Equation (\ref{Eq_L1_computation01}), corresponding to the L1 norm when considering an increasing sequence of $k = 1,2,\dots$ and corresponding discrete values $\mu = c_{k}$ for $\mu$,
\begin{equation}
L_{1}(k) = \tfrac{1}{2 T} \bigg( k c_{k} - \underbracket[0.5pt][3pt]{ \sum_{i = 1}^{k} c_{i} }_{s(k)} \bigg)
= \tfrac{1}{2 T} \big( k c_{k} - s(k) \big)
\label{Eq_auxiliary_function_fk01}
\end{equation}
with $s(k) = \sum_{i = 1}^{k} c_{i}$.
Obviously, $L_{1}(1) = 0$. Moreover, the function $L_{1}$ is a monotonic non-decreasing function. Indeed, $2 T \, L_{1}(k+1) = (k+1) c_{k+1} - s(k+1) = k c_{k+1} + c_{k+1} - s(k) - c_{k+1} = k c_{k+1} - s(k) \ge k c_{k} - s(k) = 2 T \, L_{1}(k)$.

The successive terms are computed incrementally until $L_{1}(k^{*}) < 1$ and $L_{1}(k^{*}+1) \ge 1$, which means that the admissible value of $\mu$ lies in the interval $] c_{k^{*}}, c_{k^{*}+1} ]$. This further implies that the optimal number of strictly positive elements is $k^{*}$. We thus have to perform a linear search by computing $L_{1}(k)$ and stopping when $L_{1}(k^{*}+1) \ge 1$. If this last condition is never reached, that is, $L_{1}(m) < 1$, it means that all the $m$ elements of $\mathbf{p}$ are strictly positive and thus $k^{*} = m$.

\subsubsection{Computing the optimal probability distribution}

Once the number of positive elements $k^{*}$ is computed, the probability mass $\mathbf{p}$ can be obtained from (\ref{Eq_membership_function_general01}) in which $s(k^{*})$ is known after the evaluation of (\ref{Eq_auxiliary_function_fk01}).
The procedure is summarized as follows.
\begin{enumerate}
\item Sort and renumber the $m$ elements by increasing cost ($c_{1} \le c_{2} \le \cdots \le c_{m}$).
\item Compute sequentially $L_{1}(k)$ (Equation (\ref{Eq_auxiliary_function_fk01})) for $k = 1, \dots, k^{*}$ until $k^{*} = m$ or ($L_{1}(k^{*}) < 1$ and $L_{1}(k^{*}+1) \ge 1$) (the $\mu$ parameter is in the interval [$k^{*}, k^{*} + 1$[).
\item Compute $p_{i}$ by Equation (\ref{Eq_membership_function01}) for $i = 1, \dots, k^{*}$.
\item Set $p_{i} = 0$ for $i = (k^{*}+1), \dots, m$ if $k^{*} < m$.
\item Recover the initial numbering of the elements, that is, before executing step 1 (sorting).
\end{enumerate}

\subsection{Results for the general $r>1$ case and uniform reference probabilities}
\label{Appendix_B2}

The case $r \ne 2$ and $r>1$ is a bit more complex,
\begin{equation}
\vline\,\begin{array}{llll}
\minimize\limits_{\mathbf{p}} &  \mathbf{c}^{\mathrm{T}} \mathbf{p}  +   \varUpsilon \, ( \| \mathbf{p}  \|_{r} )^{r}    \\[0.3cm]
\subjectto & \mathbf{e}^{\mathrm{T}} \mathbf{p} = 1 \\
           & \mathbf{p} \ge \mathbf{0}
\end{array}
\label{Eq_r_gt_1_problem01}
\end{equation}
where $\| . \|_{r}$ is the standard $r$-norm and we define $\varUpsilon = T / (r-1)$ for convenience.
This section proceeds similarly to the previous one and is based again on \cite{Miyamoto-1998,Kanzawa-2013}. As before, we assume that the costs $c_{i}$ and the probabilities $p_{i}$ are sorted by increasing value of cost.

\subsubsection{The Karush-Kuhn-Tucker conditions}

By proceeding as in the previous Subsection \ref{Appendix_B1}, the Karush-Kuhn-Tucker conditions are now
\begin{equation}
\begin{cases}
\mathbf{c} + r \varUpsilon \mathbf{p}^{(r-1)} - \mu \mathbf{e} - \boldsymbol{\lambda} = \mathbf{0} \\
\boldsymbol{\lambda}^{\mathrm{T}} \mathbf{p} = 0\\
\boldsymbol{\lambda} \ge \mathbf{0}
\end{cases}
\label{Eq_KKT_binary01}
\end{equation}
where $(r)$ is the elementwise $r$-power.
As before, $\lambda_{i} p_{i} = 0$ for each $i$. A reasoning similar to the $r=2$ case provides the following extension of Equation (\ref{Eq_KKT01})
\begin{equation}
p_{i} =
\begin{cases}
\sqrt[(r-1)]{ \tfrac{1}{r \varUpsilon} (\mu - c_{i}) } & \text{\small{when }} \mu - c_{i} > 0 \\
0 & \text{\small{when }} \mu - c_{i} \le 0
\end{cases}
\label{Eq_KKT_result_binary01}
\end{equation}
However, in this case, the Lagrange parameter $\mu$ is difficult to find analytically so that the procedure derived in \ref{Appendix_B1} cannot be used. Kanzawa \cite{Kanzawa-2013} therefore proposed to use a bisection method on the real line instead.

\subsubsection{A bisection procedure}

As for the $r=2$ case, the main idea is to find numerically the optimal $\mu$ by seeking the value which exactly satisfies the sum-to-one constraint $\mathbf{e}^{\mathrm{T}} \mathbf{p} = 1$. Indeed, we observe from Equation (\ref{Eq_KKT_result_binary01}) that each $p_{i}$ taken independently is strictly increasing with respect to the $\mu$ parameter when starting from the value $\mu = c_{i}$ (thus in the interval $[ c_{i}, \infty [$) \cite{Kanzawa-2013}. Moreover, from Equation (\ref{Eq_KKT_result_binary01}), $p_{i} \ge 0$. This implies that the L1 norm of vector $\mathbf{p}$ is strictly increasing from $\mu = c_{1}$ (the minimum cost), and thus a bisection method (see, e.g., \cite{Press-2007}) can be used in order to efficiently approximate this quantity. The procedure is stopped when the L1 norm is sufficiently close to $1$.
Then, once $\mu$ is closely approximated, Equation (\ref{Eq_KKT_result_binary01}) is used in order to compute the probability mass $\mathbf{p}$.

Note that the author of \cite{Kanzawa-2013} proposes, as initial lower and upper bounds for the admissible $\mu$,
\begin{equation}
\begin{cases}
\mu_{\mathrm{inf}} &= c_{1} \\
\mu_{\mathrm{sup}} &= c_{\mathrm{max}} + \dfrac{r \varUpsilon} {m^{r-1}}
\end{cases}
\label{Eq_lower_upper_bound01}
\end{equation}
where $c_{\mathrm{max}} = \max_{ i \in \{ 1, \dots, m \} } \{ c_{i} \} = c_{m}$ because $\mathbf{c}$ is sorted and $m$ is the last element of the vector.
The lower bound is obvious. However, the upper bound might need a word of explanation. Let us show that this $\mu_{\mathrm{sup}}$ necessarily leads to a L1 norm of the corresponding $\mathbf{p}$ vector greater or equal to 1.
First, Equations (\ref{Eq_KKT_result_binary01}) and (\ref{Eq_lower_upper_bound01}) imply that each of the $m$ elements of $\mathbf{p}$ is strictly positive when using $\mu_{\mathrm{sup}}$.
Then, the second expression in (\ref{Eq_lower_upper_bound01}) can be rearranged as
\begin{equation}
 \frac{(\mu_{\mathrm{sup}} - c_{\mathrm{max}})}{r \varUpsilon} = \dfrac{1} {m^{r-1}} \nonumber
\end{equation}
which implies $\sqrt[(r-1)]{ \tfrac{1}{r \varUpsilon} (\mu_{\mathrm{sup}} - c_{\mathrm{max}}) } = 1/m$. Therefore, we must have $\sqrt[(r-1)]{ \tfrac{1}{r \varUpsilon} (\mu_{\mathrm{sup}} - c_{i}) } \ge 1/m$ for each $i$. From (\ref{Eq_KKT_result_binary01}), summing this expression over $i = 1, \dots, m$ shows that the L1 norm of the corresponding $\mathbf{p}$ vector is larger or equal to $1$. The value $\mu_{\mathrm{sup}}$ in (\ref{Eq_lower_upper_bound01}) is therefore an upper bound for the admissible $\mu$ values.
The bisection procedure follows
\begin{enumerate}
\item Sort and renumber the $m$ elements by increasing cost ($c_{1} \le c_{2} \le \cdots \le c_{m}$).
\item Initialize the lower bound and the upper bound of the $\mu$ parameter as in Equation (\ref{Eq_lower_upper_bound01}).
\item Perform a bisection search on $\mu$ by computing $\| \mathbf{p} \|_{1}$ from Equation (\ref{Eq_KKT_result_binary01}) and testing if the result is lesser than or greater than 1 ($\| \mathbf{p} \|_{1}$ is strictly increasing in function of $\mu$). Stop when $\| \mathbf{p} \|_{1}$ is sufficiently close to 1.
\item Compute $\mathbf{p}$ from (\ref{Eq_KKT_result_binary01}).
\item Recover the initial numbering of the elements, that is, before executing step 1 (sorting).
\end{enumerate}

\subsection{Extension to arbitrary reference probabilities}
\label{Appendix_B3}

We now start from the objective function defined in Equation (\ref{Eq_sparse_optimization_problem_reference01}), extending (\ref{Eq_KKT_binary01}) by taking the reference probabilities into account, $\mathbf{p}_{\mathrm{ref}} = (p^{\mathrm{ref}}_{i})$ (which correspond to $\mathbf{p}_{i}^{\mathrm{ref}}$ in (\ref{Eq_sparse_optimization_problem_reference01})). The Karush-Kuhn-Tucker conditions become
\begin{equation}
\begin{cases}
\mathbf{c} + r \varUpsilon (\mathbf{p} \div \mathbf{p}_{\mathrm{ref}})^{(r-1)} - \mu \mathbf{e} - \boldsymbol{\lambda} = \mathbf{0} \\
\boldsymbol{\lambda}^{\mathrm{T}} \mathbf{p} = 0\\
\boldsymbol{\lambda} \ge \mathbf{0}
\end{cases}
\label{Eq_KKT_binary_reference01}
\end{equation}
where $\varUpsilon = T/(r-1)$ and $\div$ is the elementwise division. By a reasoning similar to the previous subsections  (\ref{Appendix_B1}-\ref{Appendix_B2}), this leads to
\begin{equation}
p_{i} =
\begin{cases}
p^{\mathrm{ref}}_{i}
\sqrt[(r-1)]{ \tfrac{1}{r \varUpsilon} (\mu - c_{i}) } & \text{\small{when }}  \mu - c_{i} > 0  \\
0 & \text{\small{when }} \mu - c_{i} \le 0
\end{cases}
\label{Eq_KKT_result_binary_reference01}
\end{equation}
Notice that in this case, the elements are still sorted by increasing cost value, but then the $p_{i}$ are no more ordered by decreasing value because they are modulated by $p_{i}^{\mathrm{ref}}$.

As in Subsection \ref{Appendix_B2}, a bisection procedure can be used in order to find the probability distribution $\mathbf{p}$ summing to one. Following the derivation in Subsection \ref{Appendix_B2}, the upper bound for the bisection procedure can be chosen here as $\mu_{\mathrm{sup}} = c_{\mathrm{max}} + r \varUpsilon$ because the reference probabilities $p_{i}^{\mathrm{ref}}$ sum to one. Thus the procedure for computing the probability distribution is exactly the same as in the previous subsection, except that Equation (\ref{Eq_KKT_result_binary_reference01}) is used instead of (\ref{Eq_KKT_result_binary01}).

\begin{center}
\rule{2.5in}{0.01in}
\end{center}

\bibliographystyle{abbrv}
\bibliography{Biblio.bib}

\ifdefined\url\else\newcommand\url[1]{\texttt{#1}}\fi
\begin{thebibliography}{100}

\bibitem{Ahuja-1993}
R.~K. Ahuja, T.~L. Magnanti, and J.~B. Orlin.
\newblock {\em Network flows: theory, algorithms, and applications}.
\newblock Prentice Hall, 1993.

\bibitem{Akamatsu-1996}
T.~Akamatsu.
\newblock Cyclic flows, {M}arkov process and stochastic traffic assignment.
\newblock {\em Transportation Research B}, 30(5):369--386, 1996.

\bibitem{Akamatsu-1997}
T.~Akamatsu.
\newblock Decomposition of path choice entropy in general transport networks.
\newblock {\em Transportation Science}, 31(4):349--362, 1997.

\bibitem{vonLuxburg-2011}
M.~Alamgir and U.~von Luxburg.
\newblock Phase transition in the family of p-resistances.
\newblock In {\em Advances in Neural Information Processing Systems 24:
  Proceedings of the NIPS 2011 conference}, pages 379--387. MIT Press, 2011.

\bibitem{Arrow-1958}
K.~Arrow, L.~Hurwicz, and H.~Uzawa.
\newblock {\em Studies in linear and non-linear programming}.
\newblock Stanford University Press, 1958.

\bibitem{Barabasi-2015}
A.~L. Barabasi.
\newblock {\em Network science}.
\newblock Cambridge University Press, 2016.

\bibitem{Bavaud-2012}
F.~Bavaud and G.~Guex.
\newblock Interpolating between random walks and shortest paths: {A} path
  functional approach.
\newblock In K.~Aberer, A.~Flache, W.~Jager, L.~Liu, J.~Tang, and C.~Gu\'eret,
  editors, {\em Proceedings of the 4th International Conference on Social
  Informatics (SocInfo '12)}, volume 7710 of {\em Lecture Notes in Computer
  Science}, pages 68--81. Springer, 2012.

\bibitem{Beck-2014}
A.~Beck.
\newblock {\em Introduction to nonlinear optimization}.
\newblock SIAM, 2014.

\bibitem{Bertsekas-1999}
D.~P. Bertsekas.
\newblock {\em Nonlinear programming}.
\newblock Athena Scientific, 2nd edition, 1999.

\bibitem{Blum-1973}
M.~Blum, R.~W. Floyd, V.~R. Pratt, R.~L. Rivest, and R.~E. Tarjan.
\newblock Time bounds for selection.
\newblock {\em Journal of Computer and System Sciences}, 7(4):448--461, 1973.

\bibitem{Borg-1997}
I.~Borg and P.~Groenen.
\newblock {\em Modern multidimensional scaling: {T}heory and applications}.
\newblock Springer, 1997.

\bibitem{Brandes-2005}
U.~Brandes and T.~Erlebach, editors.
\newblock {\em Network analysis: {M}ethodological foundations}.
\newblock Springer, 2005.

\bibitem{Brandes-2005b}
U.~Brandes and D.~Fleischer.
\newblock Centrality measures based on current flow.
\newblock In {\em Proceedings of the 22nd Annual Symposium on Theoretical
  Aspects of Computer Science (STACS '05)}, pages 533--544, 2005.

\bibitem{Buhlmann-2011}
P.~Buhlmann and S.~{van de Geer}.
\newblock {\em Statistics for high-dimensional data}.
\newblock Springer, 2011.

\bibitem{Busic-2018}
A.~Busic and S.~Meyn.
\newblock Action-constrained {M}arkov decision processes with kullback-leibler
  cost.
\newblock In {\em Proceedings of the 31st Conference On Learning Theory
  (COLT)}, pages 1431--1444. PMLR 75, 2018.

\bibitem{Chebotarev-2011}
P.~Chebotarev.
\newblock A class of graph-geodetic distances generalizing the shortest-path
  and the resistance distances.
\newblock {\em Discrete Applied Mathematics}, 159(5):295--302, 2011.

\bibitem{Chebotarev-2012}
P.~Chebotarev.
\newblock The walk distances in graphs.
\newblock {\em Discrete Applied Mathematics}, 160(10--11):1484--1500, 2012.

\bibitem{Chebotarev-2013}
P.~Chebotarev.
\newblock Studying new classes of graph metrics.
\newblock In F.~Nielsen and F.~Barbaresco, editors, {\em Proceedings of the 1st
  International Conference on Geometric Science of Information (GSI '13)},
  volume 8085 of {\em Lecture Notes in Computer Science}, pages 207--214.
  Springer, 2013.

\bibitem{Chebotarev-1997}
P.~Chebotarev and E.~Shamis.
\newblock The matrix-forest theorem and measuring relations in small social
  groups.
\newblock {\em Automation and Remote Control}, 58(9):1505--1514, 1997.

\bibitem{Chebotarev-1998a}
P.~Chebotarev and E.~Shamis.
\newblock On proximity measures for graph vertices.
\newblock {\em Automation and Remote Control}, 59(10):1443--1459, 1998.

\bibitem{chung06}
F.~Chung and L.~Lu.
\newblock {\em Complex graphs and networks}.
\newblock American Mathematical Society, 2006.

\bibitem{Condat-2016}
L.~Condat.
\newblock Fast projection onto the simplex and the $\ell_{1}$ ball.
\newblock {\em Mathematical Programming}, 158(1-2):575--585, 2016.

\bibitem{Cormen-2009}
T.~Cormen, C.~Leiserson, R.~Rivest, and C.~Stein.
\newblock {\em Introduction to algorithms}.
\newblock MIT Press, 3rd edition, 2009.

\bibitem{Cover-2006}
T.~Cover and J.~Thomas.
\newblock {\em Elements of information theory}.
\newblock Wiley, 2nd edition, 2006.

\bibitem{Culioli-2012}
J.~Culioli.
\newblock {\em Introduction a l'optimisation}.
\newblock Ellipses, 2012.

\bibitem{Delvenne-2011}
J.-C. Delvenne and A.-S. Libert.
\newblock Centrality measures and thermodynamic formalism for complex networks.
\newblock {\em Physical Review E}, 83(4):046117, 2011.

\bibitem{Demvsar-2006}
J.~Dem{\v{s}}ar.
\newblock Statistical comparisons of classifiers over multiple data sets.
\newblock {\em Journal of Machine learning research}, 7(Jan):1--30, 2006.

\bibitem{Dolan-1993}
A.~Dolan and J.~Aldous.
\newblock {\em Networks and algorithms: An introductory approach}.
\newblock Wiley, 1993.

\bibitem{Snell-1984}
P.~G. Doyle and J.~L. Snell.
\newblock {\em Random walks and electric networks}.
\newblock The Mathematical Association of America, 1984.

\bibitem{Duchi-2008}
J.~Duchi, S.~{Shalev-Shwartz}, Y.~Singer, and T.~Chandra.
\newblock Efficient projections onto the l1-ball for learning in high
  dimensions.
\newblock In {\em Proceedings of the 25th international conference on Machine
  learning (ICML '2008)}, pages 272--279, 2008.

\bibitem{Estrada-2012}
E.~Estrada.
\newblock {\em The structure of complex networks}.
\newblock Oxford University Press, 2012.

\bibitem{Estrada-2008}
E.~Estrada and N.~Hatano.
\newblock Communicability in complex networks.
\newblock {\em Physical Review E}, 77(3):036111, 2008.

\bibitem{FoussKDE-2005}
F.~Fouss, A.~Pirotte, J.-M. Renders, and M.~Saerens.
\newblock Random-walk computation of similarities between nodes of a graph,
  with application to collaborative recommendation.
\newblock {\em IEEE Transactions on Knowledge and Data Engineering},
  19(3):355--369, 2007.

\bibitem{Fouss-2016}
F.~Fouss, M.~Saerens, and M.~Shimbo.
\newblock {\em Algorithms and models for network data and link analysis}.
\newblock Cambridge University Press, 2016.

\bibitem{Fox-2016}
R.~Fox, A.~Pakman, and N.~Tishby.
\newblock G-learning: taming the noise in reinforcement learning via soft
  updates.
\newblock In {\em Proceedings of the 22nd Conference on Uncertainty in
  Artificial Intelligence (UAI 2016)}, pages 202--211, 2001.

\bibitem{Francoisse-2017}
K.~Francoisse, I.~Kivimaki, A.~Mantrach, F.~Rossi, and M.~Saerens.
\newblock A bag-of-paths framework for network data analysis.
\newblock {\em Neural Networks}, 90:90--111, 2017.

\bibitem{Fred-2003}
A.~L. Fred and A.~K. Jain.
\newblock Robust data clustering.
\newblock In {\em Proceedings of the 2003 IEEE International Computer Society
  Conference on Computer Vision and Pattern Recognition (CVPR'03)}, volume~2,
  pages 128--133, 2003.

\bibitem{Freeman-1977}
L.~C. Freeman.
\newblock A set of measures of centrality based on betweenness.
\newblock {\em Sociometry}, 40(1):35--41, 1977.

\bibitem{Garcia-Diez-2011}
S.~Garc\'{\i}a-{D}\'{\i}ez, F.~Fouss, M.~Shimbo, and M.~Saerens.
\newblock A sum-over-paths extension of edit distances accounting for all
  sequence alignments.
\newblock {\em Pattern Recognition}, 44(6):1172--1182, 2011.

\bibitem{Garcia-Diez-2011b}
S.~Garc\'{\i}a-{D}\'{\i}ez, E.~Vandenbussche, and M.~Saerens.
\newblock A continuous-state version of discrete randomized shortest-paths.
\newblock In {\em Proceedings of the 50th IEEE International Conference on
  Decision and Control (CDC '11)}, pages 6570--6577, 2011.

\bibitem{Geist-2019}
M.~Geist, B.~Scherrer, and O.~Pietquin.
\newblock A theory of regularized markov decision processes.
\newblock In {\em Proceedings of the International Conference on Machine
  Learning (ICML 2019)}, pages 2160--2169, 2019.

\bibitem{Newman2002}
M.~Girvan and M.~E.~J. Newman.
\newblock Community structure in social and biological networks.
\newblock {\em Proceedings of the National Academy of Sciences of the USA},
  99(12):7821--7826, 2002.

\bibitem{Griva-2008}
I.~Griva, S.~Nash, and A.~Sofer.
\newblock {\em Linear and nonlinear optimization}.
\newblock SIAM, 2nd edition, 2008.

\bibitem{Guex-2015}
G.~Guex and F.~Bavaud.
\newblock Flow-based dissimilarities: shortest path, commute time, max-flow and
  free energy.
\newblock In B.~Lausen, S.~{Krolak-Schwerdt}, and M.~Bohmer, editors, {\em Data
  science, learning by latent structures, and knowledge discovery}, volume 1564
  of {\em Studies in Classification, Data Analysis, and Knowledge
  Organization}, pages 101--111. Springer, 2015.

\bibitem{Guex-2019Cov}
G.~Guex, S.~Courtain, and M.~Saerens.
\newblock Covariance and correlation kernels on a graph in the generalized
  bag-of-paths formalism.
\newblock {\em arXiv preprint arXiv:1902.03002 submitted for publication},
  2019.

\bibitem{Guex-2019}
G.~Guex, I.~Kivimaki, and M.~Saerens.
\newblock Randomized optimal transport on a graph: framework and new distance
  measures.
\newblock {\em Network Science}, 7(1):88--122, 2019.

\bibitem{Hashimoto-2015}
T.~Hashimoto, Y.~Sun, and T.~Jaakkola.
\newblock From random walks to distances on unweighted graphs.
\newblock In {\em Advances in Neural Information Processing Systems 24:
  Proceedings of the NIPS '15 Conference}, 2015.

\bibitem{Hastie-2009}
T.~Hastie, R.~Tibshirani, and J.~Friedman.
\newblock {\em The elements of statistical learning: {D}ata mining, inference,
  and prediction}.
\newblock Springer, 2009.

\bibitem{Hastie-2015}
T.~Hastie, R.~Tibshirani, and M.~Wainwright.
\newblock {\em Statistical learning with sparsity}.
\newblock CRC Press, 2015.

\bibitem{Havrda-1967}
H.~Havrda and F.~Charvat.
\newblock Quantification method of classification processes. concept of
  structural $\alpha$-entropy.
\newblock {\em Kybernetika}, 3(1):30–--35, 1967.

\bibitem{Hazan-2007}
T.~Hazan, R.~Hardoon, and A.~Shashua.
\newblock Plsa for sparse arrays with {Tsallis} pseudo-additive divergence:
  noise robustness and algorithm.
\newblock In {\em Proceedings of the 11th IEEE International Conference on
  Computer Vision}, pages 1--8. IEEE, 2007.

\bibitem{Hazan-2007b}
T.~Hazan and A.~Shashua.
\newblock An efficient algorithm for maximum {Tsallis} entropy using
  fenchel-duality.
\newblock Technical Report TR-110, The Hebrew University of Jerusalem, Israel,
  2007.

\bibitem{Herbster-2009}
M.~Herbster and G.~Lever.
\newblock Predicting the labelling of a graph via minimum p-seminorm
  interpolation.
\newblock In {\em Proceedings of the 22nd Conference on Learning Theory (COLT
  '09)}, pages 18--21, 2009.

\bibitem{Hubert-1985}
L.~Hubert and P.~Arabie.
\newblock Comparing partitions.
\newblock {\em Journal of classification}, 2(1):193--218, 1985.

\bibitem{ivashkin-2016}
V.~Ivashkin and P.~Chebotarev.
\newblock Do logarithmic proximity measures outperform plain ones in graph
  clustering?
\newblock In {\em International Conference on Network Analysis}, pages 87--105.
  Springer, 2016.

\bibitem{Jaynes-1957}
E.~T. Jaynes.
\newblock Information theory and statistical mechanics.
\newblock {\em Physical Review}, 106:620--630, 1957.

\bibitem{Kanzawa-2013}
Y.~Kanzawa.
\newblock Generalization of quadratic regularized and standard fuzzy c-means
  clustering with respect to regularization of hard c-means.
\newblock In V.~Torra, Y.~Narukawa, G.~Navarro-Arribas, and D.~Meg{\'i}as,
  editors, {\em Modeling Decisions for Artificial Intelligence}, pages
  152--165, Berlin, Heidelberg, 2013. Springer Berlin Heidelberg.

\bibitem{Kanzawa-2018}
Y.~Kanzawa.
\newblock Q-divergence-based relational fuzzy c-means clustering.
\newblock {\em Journal of Advanced Computational Intelligence and Intelligent
  Informatics}, 22(1):34--43, 2018.

\bibitem{Kappen-2012}
H.~J. Kappen, V.~G{\'o}mez, and M.~Opper.
\newblock Optimal control as a graphical model inference problem.
\newblock {\em Machine learning}, 87(2):159--182, 2012.

\bibitem{Kapur-1989}
J.~N. Kapur.
\newblock {\em Maximum-entropy models in science and engineering}.
\newblock Wiley, 1989.

\bibitem{Katz-1953}
L.~Katz.
\newblock A new status index derived from sociometric analysis.
\newblock {\em Psychometrika}, 18(1):39--43, 1953.

\bibitem{Keylock-2005}
C.~Keylock.
\newblock Simpson diversity and the shannon-wiener index as special cases of a
  generalized entropy.
\newblock {\em Oikos}, 109(1):203--207, 2005.

\bibitem{Kivimaki-2012}
I.~Kivim{\"a}ki, M.~Shimbo, and M.~Saerens.
\newblock Developments in the theory of randomized shortest paths with a
  comparison of graph node distances.
\newblock {\em Physica A: Statistical Mechanics and its Applications},
  393:600--616, 2014.

\bibitem{Klein-1993}
D.~J. Klein and M.~Randic.
\newblock Resistance distance.
\newblock {\em Journal of Mathematical Chemistry}, 12(1):81--95, 1993.

\bibitem{Kolaczyk-2009}
E.~D. Kolaczyk.
\newblock {\em Statistical analysis of network data: {M}ethods and models}.
\newblock Springer Series in Statistics. Springer, 2009.

\bibitem{Kondor-2002}
R.~I. Kondor and J.~Lafferty.
\newblock Diffusion kernels on graphs and other discrete structures.
\newblock In {\em Proceedings of the 19th International Conference on Machine
  Learning (ICML '02)}, pages 315--322, 2002.

\bibitem{Laha-2018}
A.~Laha, S.~A. Chemmengath, P.~Agrawal, M.~Khapra, K.~Sankaranarayanan, and
  H.~Ramaswamy.
\newblock On controllable sparse alternatives to softmax.
\newblock In {\em Advances in Neural Information Processing Systems 32:
  Proceedings of the NeurIPS '18 Conference}, pages 6422--6432, 2018.

\bibitem{lancichinetti-2008}
A.~Lancichinetti, S.~Fortunato, and F.~Radicchi.
\newblock Benchmark graphs for testing community detection algorithms.
\newblock {\em Physical review E}, 78(4):046110, 2008.

\bibitem{Lang-1995}
K.~Lang.
\newblock Newsweeder: Learning to filter netnews.
\newblock In {\em Proceedings of the 12th International Machine Learning
  Conference (ML95)}, pages 331--339, 1995.

\bibitem{Lee-2018b}
K.~Lee, S.~Choi, and S.~Oh.
\newblock Maximum causal {Tsallis} entropy imitation learning.
\newblock In {\em Advances in Neural Information Processing Systems 31:
  Proceedings of the NIPS 2010 Conference}, pages 4403--4413, 2018.

\bibitem{Lee-2018}
K.~{Lee}, S.~{Choi}, and S.~{Oh}.
\newblock Sparse {Markov} decision processes with causal sparse {Tsallis}
  entropy regularization for reinforcement learning.
\newblock {\em IEEE Robotics and Automation Letters}, 3(3):1466--1473, 2018.

\bibitem{Lewis-2009}
T.~Lewis.
\newblock {\em Network science}.
\newblock Wiley, 2009.

\bibitem{Li-2011}
Y.~Li, Z.-L. Zhang, and D.~Boley.
\newblock The routing continuum from shortest-path to all-path: {A} unifying
  theory.
\newblock In {\em Proceedings of the 31st International Conference on
  Distributed Computing Systems (ICDCS '11)}, pages 847--856. IEEE Computer
  Society, 2011.

\bibitem{Li-2013}
Y.~Li, Z.-L. Zhang, and D.~Boley.
\newblock From shortest-path to all-path: {T}he routing continuum theory and
  its applications.
\newblock {\em IEEE Transactions on Parallel and Distributed Systems},
  25(7):1745--1755, 2013.

\bibitem{Luenberger-1979}
D.~G. Luenberger.
\newblock {\em Introduction to dynamic systems: {T}heory, models, and
  applications}.
\newblock Wiley, 1979.

\bibitem{Luenberger-2010}
D.~G. Luenberger and Y.~Ye.
\newblock {\em Linear and nonlinear programming}.
\newblock Springer, 3rd edition, 2010.

\bibitem{Lusseau-2003-emergent}
D.~Lusseau.
\newblock The emergent properties of a dolphin social network.
\newblock {\em Proceedings of the Royal Society of London. Series B: Biological
  Sciences}, 270(suppl\_2):S186--S188, 2003.

\bibitem{Lusseau-2003-bottlenose}
D.~Lusseau, K.~Schneider, O.~J. Boisseau, P.~Haase, E.~Slooten, and S.~M.
  Dawson.
\newblock The bottlenose dolphin community of doubtful sound features a large
  proportion of long-lasting associations.
\newblock {\em Behavioral Ecology and Sociobiology}, 54(4):396--405, 2003.

\bibitem{Macskassy-07}
S.~A. Macskassy and F.~Provost.
\newblock Classification in networked data: {A} toolkit and a univariate case
  study.
\newblock {\em Journal of Machine Learning Research}, 8:935--983, 2007.

\bibitem{Manning-2008}
C.~Manning, P.~Raghavan, and H.~Sch\"utze.
\newblock {\em Introduction to information retrieval}.
\newblock Cambridge University Press, 2008.

\bibitem{Martins-2016}
A.~Martins and R.~Astudillo.
\newblock From softmax to sparsemax: a sparse model of attention and
  multi-label classification.
\newblock In {\em Proceedings of the International Conference on Machine
  Learning (ICML-2016)}, pages 1614--1623, 2016.

\bibitem{Menard-2003}
M.~Menard, V.~Courboulay, and P.-A. Dardignac.
\newblock Possibilistic and probabilistic fuzzy clustering: unification within
  the framework of the non-extensive thermostatistics.
\newblock {\em Pattern Recognition}, 36(6):1325--1342, 2003.

\bibitem{Minoux-1986}
M.~Minoux.
\newblock {\em Mathematical programming, theory and algorithms}.
\newblock John Wiley, 1986.

\bibitem{Miyamoto-2008}
S.~Miyamoto, H.~Ichihashi, and K.~Honda.
\newblock {\em Algorithms for fuzzy clustering}.
\newblock Springer, 2008.

\bibitem{Miyamoto-1998}
S.~Miyamoto and K.~Umayahara.
\newblock Fuzzy clustering by quadratic regularization.
\newblock In {\em Proceedings of the IEEE International Conference on Fuzzy
  Systems}, pages 1394--1399, 1998.

\bibitem{Muzellec-2017}
B.~Muzellec, R.~Nock, G.~Patrini, and F.~Nielsen.
\newblock Tsallis regularized optimal transport and ecological inference.
\newblock In {\em Proceedings of the 31 International Conference of the
  Association for the Advancement of Artificial Intelligence (AAAI 2017)},
  2017.

\bibitem{Newman-05}
M.~E.~J. Newman.
\newblock A measure of betweenness centrality based on random walks.
\newblock {\em Social Networks}, 27(1):39--54, 2005.

\bibitem{Newman-2018}
M.~E.~J. Newman.
\newblock {\em Networks: {A}n introduction, 2nd ed}.
\newblock Oxford University Press, 2018.

\bibitem{Nguyen-2016}
C.~Ngyen and H.~Mamitsuka.
\newblock New resistance distances with global information on large graphs.
\newblock In {\em Proceedings of the 19th International Conference on
  Artificial Intelligence and Statistics (AISTATS '16)}, pages 639--647, 2016.

\bibitem{Norris-1997}
J.~R. Norris.
\newblock {\em Markov chains}.
\newblock Cambridge University Press, 1997.

\bibitem{Panzacchi-2016}
M.~Panzacchi, B.~{Van Moorter}, O.~Strand, M.~Saerens, I.~Kivimaki, C.~{St
  Clair}, I.~Herfindal, and L.~Boitani.
\newblock Predicting the continuum between corridors and barriers to animal
  movements using step selection functions and randomized shortest paths.
\newblock {\em Journal of Animal Ecology}, 85(1):32--42, 2016.

\bibitem{Peliti-2011}
L.~Peliti.
\newblock {\em Statistical mechanics in a nutshell}.
\newblock Princeton University Press, 2011.

\bibitem{Press-2007}
W.~Press, S.~Teukolsky, W.~Vetterling, and B.~Flannery.
\newblock {\em Numerical recipes: {T}he art of scientific computing}.
\newblock Cambridge University Press, 3rd edition, 2007.

\bibitem{Price-1971}
W.~L. Price.
\newblock {\em Graphs and networks: an introduction}.
\newblock London Butterworths, 1971.

\bibitem{Rand-1971}
W.~M. Rand.
\newblock Objective criteria for the evaluation of clustering methods.
\newblock {\em Journal of the American Statistical association},
  66(336):846--850, 1971.

\bibitem{Rardin-1998}
R.~Rardin.
\newblock {\em Optimization in operations research}.
\newblock Prentice Hall, 1998.

\bibitem{Reichl-1998}
L.~E. Reichl.
\newblock {\em A modern course in statistical physics}.
\newblock Wiley, 2nd edition, 1998.

\bibitem{Rubin-2012}
J.~Rubin, O.~Shamir, and N.~Tishby.
\newblock {\em Trading value and information in MDPs}, pages 57--74.
\newblock Springer Berlin Heidelberg, Berlin, Heidelberg, 2012.

\bibitem{Saerens-2008}
M.~Saerens, Y.~Achbany, F.~Fouss, and L.~Yen.
\newblock Randomized shortest-path problems: {T}wo related models.
\newblock {\em Neural Computation}, 21(8):2363--2404, 2009.

\bibitem{Scholkopf-2002}
B.~Sch\"olkopf and A.~Smola.
\newblock {\em Learning with kernels}.
\newblock MIT Press, 2002.

\bibitem{Silva-2016}
T.~Silva and L.~Zhao.
\newblock {\em Machine learning in complex networks}.
\newblock Springer, 2016.

\bibitem{Sommer-2016}
F.~Sommer, F.~Fouss, and M.~Saerens.
\newblock Comparison of graph node distances on clustering tasks.
\newblock In {\em Proceedings of the International Conference on Artificial
  Neural Networks (ICANN 2016). Lecture Notes in Computer Science}, volume
  9886, pages 192--201, 2016.
\newblock Springer.

\bibitem{Sommer-2017}
F.~Sommer, F.~Fouss, and M.~Saerens.
\newblock Modularity-driven kernel k-means for community detection.
\newblock In {\em Proceedings of the International Conference on Artificial
  Neural Networks (ICANN 2017). Lecture Notes in Computer Science}, volume
  10614, pages 423--433, 2017.
\newblock Springer.

\bibitem{Strehl-2002}
A.~Strehl and J.~Ghosh.
\newblock Cluster ensembles---a knowledge reuse framework for combining
  multiple partitions.
\newblock {\em Journal of machine learning research}, 3(Dec):583--617, 2002.

\bibitem{Tang-2009}
L.~Tang and H.~Liu.
\newblock Relational learning via latent social dimensions.
\newblock In {\em Proceedings of the 15th ACM SIGKDD International Conference
  on Knowledge Discovery and Data Mining (KDD '09)}, pages 817--826, 2009.

\bibitem{Tang-2009b}
L.~Tang and H.~Liu.
\newblock Scalable learning of collective behavior based on sparse social
  dimensions.
\newblock In {\em Proceedings of the ACM Conference on Information and
  Knowledge Management (CIKM '09)}, pages 1107--1116, 2009.

\bibitem{Tang-2010}
L.~Tang and H.~Liu.
\newblock Toward predicting collective behavior via social dimension
  extraction.
\newblock {\em IEEE Intelligent Systems}, 25(4):19--25, 2010.

\bibitem{Taylor-1998}
H.~M. Taylor and S.~Karlin.
\newblock {\em An introduction to stochastic modeling}.
\newblock Academic Press, 3rd edition, 1998.

\bibitem{Taylor-1996}
P.~D. Taylor.
\newblock Inclusive fitness arguments in genetic models of behaviour.
\newblock {\em Journal of Mathematical Biology}, 34(5--6):654--674, 1996.

\bibitem{Thelwall04}
M.~Thelwall.
\newblock {\em Link analysis: {A}n information science approach}.
\newblock Elsevier, 2004.

\bibitem{Theodorou-2013}
E.~A. Theodorou, D.~Krishnamurthy, and E.~Todorov.
\newblock From information theoretic dualities to path integral and
  kullback-leibler control: continuous and discrete time formulations.
\newblock In {\em The Sixteenth Yale Workshop on Adaptive and Learning
  Systems}, 2013.

\bibitem{Theodorou-2012}
E.~A. Theodorou and E.~Todorov.
\newblock Relative entropy and free energy dualities: Connections to path
  integral and kl control.
\newblock In {\em Proceedings of the 51st IEEE Conference on Decision and
  Control (CDC 2012)}, pages 1466--1473. IEEE, 2012.

\bibitem{Todorov-2007}
E.~Todorov.
\newblock Linearly-solvable {Markov} decision problems.
\newblock In {\em Advances in Neural Information Processing Systems 19 (NIPS
  2006)}, pages 1369--1375. {MIT} Press, 2007.

\bibitem{Todorov-2008}
E.~Todorov.
\newblock General duality between optimal control and estimation.
\newblock In {\em Proceedings of 47th IEEE Conference on Decision and Control
  (CDC'08)}, pages 4286--4292, 2008.

\bibitem{Tsallis-1998}
C.~Tsallis.
\newblock Generalized entropy-based criterion for consistent testing.
\newblock {\em Physical Review E}, 58(2):1442, 1998.

\bibitem{Tsallis-2009}
C.~Tsallis.
\newblock {\em Introduction to nonextensive statistical mechanics}.
\newblock Springer, 2009.

\bibitem{Luxburg-2010}
U.~{von Luxburg}, A.~Radl, and M.~Hein.
\newblock Getting lost in space: {L}arge sample analysis of the commute
  distance.
\newblock In {\em Advances in Neural Information Processing Systems 23:
  Proceedings of the NIPS '10 Conference}, pages 2622--2630, 2010.

\bibitem{Luxburg-2014}
U.~{von Luxburg}, A.~Radl, and M.~Hein.
\newblock Hitting and commute times in large random neighborhood graphs.
\newblock {\em Journal of Machine Learning Research}, 15:1751--1798, 2014.

\bibitem{Wang-2013b}
W.~Wang and M.~Carreira-Perpinan.
\newblock Projection onto the probability simplex: an efficient algorithm with
  a simple proof, and an application.
\newblock {\em ArXiv preprint arXiv:1309.1541 [cs.LG]}, 2013.

\bibitem{Wasserman-1994}
S.~Wasserman and K.~Faust.
\newblock {\em Social network analysis: {M}ethods and applications}.
\newblock Cambridge University Press, 1994.

\bibitem{Wilcoxon-1945}
F.~Wilcoxon.
\newblock Individual comparisons by ranking methods.
\newblock {\em Biometrics Bulletin}, 1(6):80--83, 1945.

\bibitem{Yen-2007}
L.~Yen, F.~Fouss, C.~Decaestecker, P.~Francq, and M.~Saerens.
\newblock Graph nodes clustering based on the commute-time kernel.
\newblock In {\em Proceedings of the 11th Pacific-Asia Conference on Knowledge
  Discovery and Data Mining (PAKDD '07)}, volume 4426 of {\em Lecture Notes in
  Artificial Intelligence}, pages 1037--1045. Springer, 2007.

\bibitem{Yen-2009}
L.~Yen, F.~Fouss, C.~Decaestecker, P.~Francq, and M.~Saerens.
\newblock Graph nodes clustering with the sigmoid commute-time kernel: A
  comparative study.
\newblock {\em Data \& Knowledge Engineering}, 68(3):338--361, 2009.

\bibitem{Yen-08K}
L.~Yen, A.~Mantrach, M.~Shimbo, and M.~Saerens.
\newblock A family of dissimilarity measures between nodes generalizing both
  the shortest-path and the commute-time distances.
\newblock In {\em Proceedings of the 14th ACM SIGKDD International Conference
  on Knowledge Discovery and Data Mining (KDD '08)}, pages 785--793, 2008.

\bibitem{Zachary1977}
W.~W. Zachary.
\newblock An information flow model for conflict and fission in small groups.
\newblock {\em Journal of Anthropological Research}, 33(4):452--473, 1977.

\bibitem{Zhang-2008b}
D.~Zhang and R.~Mao.
\newblock Classifying networked entities with modularity kernels.
\newblock In {\em Proceedings of the 17th ACM Conference on Information and
  Knowledge Management (CIKM 2008)}, pages 113--122. ACM, 2008.

\bibitem{Zhang-2008}
D.~Zhang and R.~Mao.
\newblock A new kernel for classification of networked entities.
\newblock In {\em Proceedings of 6th International Workshop on Mining and
  Learning with Graphs}, Helsinki, Finland, 2008.

\end{thebibliography}

\begin{center}
\rule{2.5in}{0.01in}
\end{center}

\end{document}